\documentclass[10pt,journal,compsoc]{IEEEtran}

\ifCLASSOPTIONcompsoc
  \usepackage[nocompress]{cite}
\else
  \usepackage{cite}
\fi

\ifCLASSINFOpdf
  \usepackage[pdftex]{graphicx}
  \graphicspath{{../figures/}}
  \DeclareGraphicsExtensions{.pdf}
\else
  \usepackage[dvips]{graphicx}
  \graphicspath{{../figures/}}
  \DeclareGraphicsExtensions{.pdf}
\fi

\newcommand{\re}[1]{\textcolor{black}{#1}}
\newcommand{\rv}[1]{\textcolor{black}{#1}}

\usepackage{longtable}
\usepackage{listings}
\usepackage{xcolor}
\usepackage{hyperref}       
\definecolor{mygreen}{rgb}{0,0.6,0}
\definecolor{mygray}{rgb}{0.5,0.5,0.5}
\definecolor{mymauve}{rgb}{0.58,0,0.82}
\usepackage{hyperref}       
\usepackage{url}            
\usepackage{booktabs}       
\usepackage{amsfonts}       
\usepackage{nicefrac}       
\usepackage{microtype}      
\usepackage{xcolor}         

\usepackage{microtype}
\usepackage{graphicx}
\usepackage{caption}
\usepackage{subcaption}

\usepackage{booktabs} 

\usepackage{amsmath}
\usepackage{amssymb}
\usepackage{mathtools}
\usepackage{amsthm}

\usepackage{array}

\usepackage{multirow}
\usepackage{hyperref}
\usepackage{enumerate}
\usepackage{xcolor}
\usepackage{comment}
\usepackage{colortbl}

\usepackage{listings}

\usepackage[capitalize,noabbrev]{cleveref}
\newcommand{\appsimprove}{\rv{5.8\%}}
\newcommand{\codecontestsimprove}{\rv{5.9\%}}
\newcommand{\reflectimprove}{3.3\%}

\usepackage{enumitem}

\usepackage{xspace}
\newcommand{\sys}{MoTCoder\xspace}
\newcommand{\idea}{Module-of-Thought\xspace}

\begin{document}

\title{\sys: Elevating Large Language Models \\ with \idea}

\author{Jingyao~Li,
        Pengguang~Chen,
        Bin~Xia,
        Hong~Xu,
        and~Jiaya~Jia,~\IEEEmembership{Fellow,~IEEE}
\IEEEcompsocitemizethanks{\IEEEcompsocthanksitem Jingyao Li, Bin Xia and Hong Xu are from the Department of Computer Science and Engineering of the Chinese University of Hong Kong (CUHK) \\
Jiaya Jia's E-mail: leojia9@gmail.com
\IEEEcompsocthanksitem Pengguang Chen and Jiaya Jia are with SmartMore.}
\thanks{Manuscript received Aug. 24th, 2024.}}

\markboth{Journal of \LaTeX\ Class Files,~Vol.~14, No.~8, August~2015}%
{Shell \MakeLowercase{\textit{et al.}}: Bare Demo of IEEEtran.cls for Computer Society Journals}

\IEEEtitleabstractindextext{%
\begin{abstract}
Large Language Models (LLMs) have showcased impressive capabilities in handling straightforward programming tasks. However, their performance tends to falter when confronted with more challenging programming problems. We observe that conventional models often generate solutions as monolithic code blocks, restricting their effectiveness in tackling intricate questions. To overcome this limitation, we present {\idea Coder (\sys)}. We introduce a framework for MoT instruction tuning, designed to promote the decomposition of tasks into logical sub-tasks and sub-modules. Our investigations reveal that, through the cultivation and utilization of sub-modules, \sys significantly improves both the modularity and correctness of the generated solutions, leading to substantial \emph{pass@1} improvements of \appsimprove{} on APPS and \codecontestsimprove{} on CodeContests. \re{MoTCoder also achieved significant improvements in self-correction capabilities, surpassing the current SOTA by \reflectimprove{}. Additionally, we provide an analysis of between problem complexity and optimal module decomposition and evaluate the maintainability index, confirming that the code generated by MoTCoder is easier to understand and modify, which can be beneficial for long-term code maintenance and evolution.} Our codes are available at \href{https://github.com/dvlab-research/MoTCoder}{https://github.com/dvlab-research/MoTCoder}.
\end{abstract}

}

\maketitle
\IEEEdisplaynontitleabstractindextext
\IEEEpeerreviewmaketitle

\IEEEraisesectionheading{\section{Introduction}\label{sec:intro}}
\IEEEPARstart{D}eveloping systems that can generate executable and functionally correct computer programs has long been sought after in artificial intelligence~\cite{Zohar71}. 
Recently, Large Language Models (LLMs)~\cite{GPT3, GPT4, PaLM, palm2, Chinchilla, gopher, GLM-130B, llama, opt} have showcased remarkable success in many problem domains beyond natural language processing, and is poised as a promising approach to tackle code modeling and generation (as well as other coding related tasks)~\cite{codellama, black10gpt, humeval}.
Through instruction fine-tuning~\cite{starcoder, wizardcoder, wang2023codet5+, AlphaCode, codegen, CodeGeeX, incoder, humeval, codet5}, LLMs have achieved impressive performance on code generation benchmarks like HumanEval~\cite{humeval} and MBPP~\cite{MBPP}.

\begin{figure}[t]
    \centering
    \includegraphics[width=\linewidth]{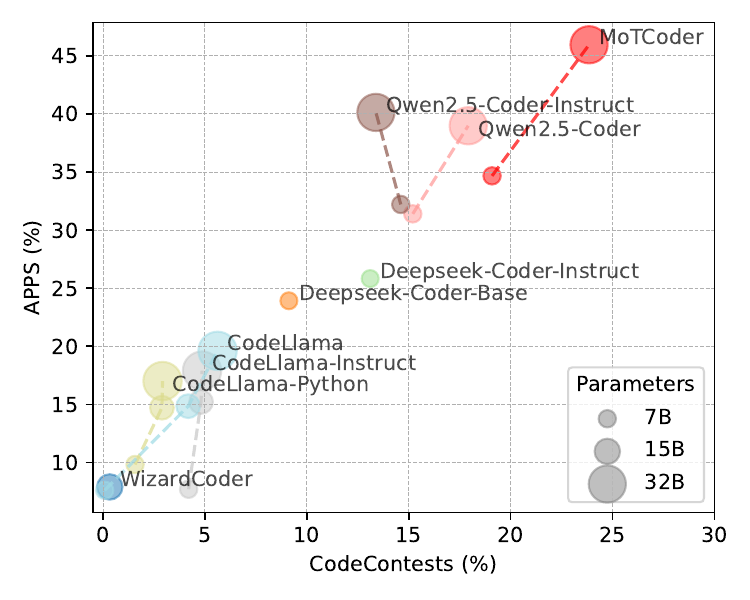}
    \captionof{figure}{\re{\emph{Pass@1} results on CodeContests (x-axis) and APPS (y-axis). Comparison of our MoTCoder with previous SOTAs. Model size are indicated by scatter size.}}
    \label{fig:impression}
\end{figure}

Yet, when confronted with intricate coding problems such as APPS \cite{hendrycksapps2021} and CodeContests ~\cite{AlphaCode}, SOTA models struggle to match seasoned developers \cite{hendrycksapps2021, AlphaCode, shinn2023reflexion}.
The main culprit is their overly simplistic generation approach. 
Current models produce the code solution as a single monolithic block in an attempt to solve the problem in one shot.
While feasible for simple tasks, this approach becomes increasingly inefficient to tackle a complex task which naturally entails solving multiple sub-tasks. 
In contrast, adept developers often devise modularized solutions by breaking down the original problem into more approachable components that can be individually and more efficiently solved. 

Following this intuition, recent works~\cite{jiang2023selfplanning, codechain} propose iterative code inference for code generation. 
They, however, come with added inference costs and offer only marginal performance gains. 
In this work, we propose to more efficiently improve the modularization capability of coding LLMs using \idea (MoT) instruction fine-tuning.  
Our approach guides LLMs to break down their solution into modular components, each representing an abstract function dedicated to a logical sub-task. To train the model to adhere to the MoT prompt, we generate instructional data using a process termed MoT Code Instruction Transformation. In this process, LLMs are instructed to outline necessary modules, generating only their function headers and docstrings that describe their intended usage. 
Subsequently, the instruction guides the model to implement these modules and eventually combine them into the final solution. After that, we fine-tune the LLM with our MoT instruction fune-tuning, resulting in our \sys model.

\re{Our experiments demonstrate that MoTCoder establishes new SOTA results on challenging code generation benchmarks such as APPS and CodeContests. Specifically, MoTCoder improves the \emph{pass@1} performance by significant margins, exemplified by improvements over existing metrics by \appsimprove{} on APPS and \codecontestsimprove{} on CodeContests, as illustrated in \cref{fig:impression}. MoTCoder also achieved significant improvements in self-correction capabilities, surpassing the current SOTA by \reflectimprove{}.}

\re{Beyond the pass@k metric, we have conducted a detailed analysis and comprehensive evaluation of MoTCoder. In \cref{sec:discuss}, we analyze the relationship between problem complexity and optimal module decomposition, confirming that finer modular decomposition is beneficial for enhancing model performance in complex problems. This helps explain why the MoT approach is more effective than traditional methods for complex programming challenges. We also perform a quantitative analysis on the time and memory usage across different problem scales, verifying that our approach can reduce memory consumption. Moreover, through the evaluation of the maintainability index, we confirm that the code generated by MoTCoder is easier to understand and modify, which can be beneficial for long-term code maintenance and evolution.}

In summary, our contributions are threefold:
\begin{enumerate}
    \item We propose a 2-step \emph{\idea Code Instruction Transformation} approach that instructs LLMs to generate modularized code solutions. 
    
    \item \re{We develop \idea Coder (\sys), a new model that enhances the modularization capabilities of LLMs with \emph{MoT Instruction Tuning}. It breaks down complex problems into sub-modules, and it has been validated to improve the model's maintainability index.}
    
    \item \re{Our approach not only achieves SOTA performance on challenging programming tasks, including APPS and CodeContests, but also demonstrates strong self-repair capabilities.}
\end{enumerate}

\begin{figure}[t]
\centering
\label{fig:instruction}
\end{figure}

\begin{figure*}[t]
\centering
\includegraphics[width=\linewidth, trim=35 135 55 47, clip]{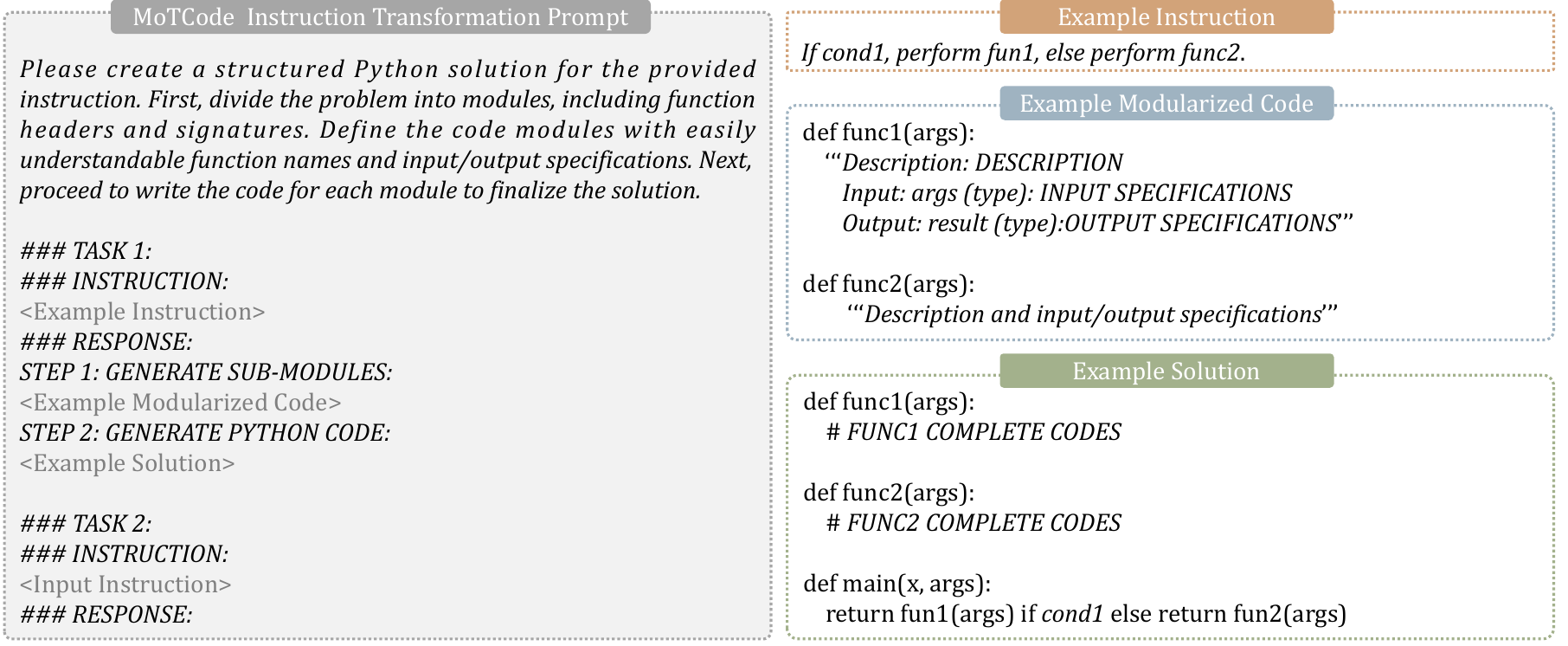}
\caption{\textbf{Left:} To transform conventional instructions into module-of-thought instructions, our MoT instruction transformation approach guides the models through a two-step process. Initially, the models are instructed to outline the necessary sub-modules, generating only their function headers and docstrings that describe their intended usage. The subsequent instruction then guides the model to implement these sub-modules and integrate them into a cohesive final solution. This instruction is complemented by a one-shot example to further enhance understanding. \textbf{Right:} Examples of instruction, modularized code and solution. }
\label{fig:prompts}
\end{figure*}

\begin{figure*}[!t]
\centering
\includegraphics[width=0.99\linewidth, trim=3 3 400 3, clip]{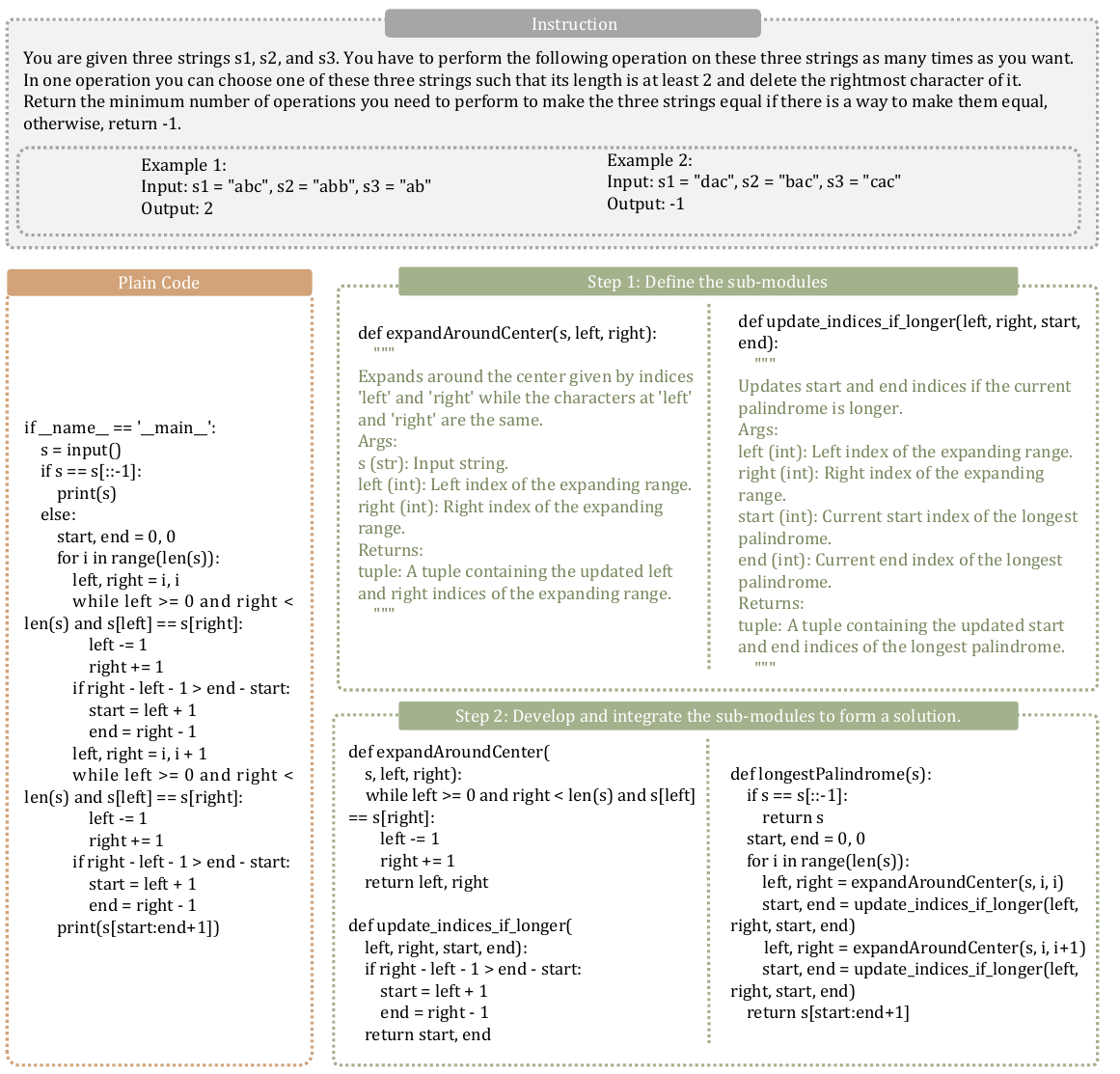}
\caption{Illustration of our 2-step Module-of-Thought Instruction Transformation: Plain code utilizes a single module to directly generate code. In contrast, our Module-of-Thought Instruction Transformation first instructs LLMs to outline necessary sub-modules, generating only their function headers and docstrings that describe their intended usage. Subsequently, the instructions guide the model in implementing these sub-modules and eventually combining them into a comprehensive final solution.}
\label{fig:framework}
\end{figure*}

\section{Related Works}\label{sec:related}
\subsection{Large Language Models}
\textbf{General LLMs.} In recent times, Large Language Models (LLMs) have exhibited remarkable prowess across a wide array of tasks. Leading technology companies have made significant advancements in developing highly proficient closed-source LLMs, including OpenAI's GPT3~\cite{GPT3} and GPT4~\cite{GPT4}, Google's PaLM~\cite{PaLM,palm2}, Bard\footnote{\scriptsize\url{https://bard.google.com/}}, DeepMind's Chinchilla~\cite{Chinchilla}, and Gopher~\cite{gopher}, as well as Anthropic's Claude\footnote{\scriptsize\url{https://www.anthropic.com/index/introducing-claude}}. The AI community has also observed the release of several open-source LLMs, where model weights are made publicly available. EleutherAI has contributed GPT-NeoX-20B~\cite{GPT-NeoX-20B} and GPT-J-6B~\cite{gpt-j}. Google has released UL2-20B~\cite{UL2}. Tsinghua University has introduced GLM-130B~\cite{GLM-130B}. Meta has released OPT~\cite{opt} and LLaMA~\cite{llama}. 
\re{Recently, the Qwen series~\cite{qwen, qwen2, qwen25} and DeepSeek series~\cite{deepseek, deepseekv2, deepseekv3} models have emerged as significant contributors, achieving SOTA performance across a variety of benchmarks. }

\textbf{Coding LLMs.} Recent research has introduced a significant number of LLMs tailored for code-related tasks to address the challenges of code understanding and generation. Closed-source models include OpenAI's Codex~\cite{humeval} and Code-Davinci~\cite{Azure}. Google has proposed PaLM-Coder~\cite{PaLM}. These models excel on popular code completion benchmarks such as HumanEval~\cite{humeval} and MBPP~\cite{MBPP}. On the open-source front, Salesforce has introduced CodeGen~\cite{codegen}, CodeT5~\cite{codet5}, and CodeT5+~\cite{CodeT5+}. Tsinghua University has contributed CodeGeeX~\cite{CodeGeeX}, and the BigCode Project has developed StarCoder~\cite{starcoder}. \re{Furthermore, the latest DeepSeek-Coder series~\cite{deepseekcoder, deepseekv2} and Qwen-Coder~\cite{qwen25coder} models have set new standards by achieving SOTA results on multiple coding benchmarks. }

\subsection{Instruction Fine-Tuning}
\textbf{General Instruction Tuning.} In its early stages, the core aim of instruction fine-tuning was to amplify the cross-task generalization capabilities of Language Models (LMs). This was accomplished by subjecting LMs to fine-tuning using an extensive corpus of public Natural Language Processing (NLP) tasks. Pioneering this approach, T5~\cite{t5} underwent training on a diverse set of supervised text-to-text tasks. Subsequent endeavors like FLAN~\cite{DBLP:conf/iclr/WeiBZGYLDDL22}, ExT5~\cite{ExT5}, T0~\cite{T0}, and UnifiedQA~\cite{UnifiedQA} broadened the spectrum of tasks, fortifying the overall generalization capability of LMs. Noteworthy contributions from ZeroPrompt~\cite{ZeroPrompt} and FLAN-T5~\cite{flan-t5} pushed boundaries by incorporating thousands of tasks into their training pipelines. OpenAI has taken an alternative route by enlisting human annotators to contribute an extensive corpus of human instructions, encompassing diverse formats and a broad spectrum of task types. Building upon this dataset, OpenAI trained its GPT-3~\cite{GPT3} model to create InstructGPT~\cite{DBLP:conf/nips/Ouyang0JAWMZASR22}, which better aligns with users' inputs. This developmental trajectory has given rise to notable works such as ChatGPT. In the open-source realm, Alpaca~\cite{alpaca} adopts the self-instruct method~\cite{wang2022self}, leveraging ChatGPT to generate data for training. Vicuna~\cite{vicuna2023} utilizes user-shared conversations collected from ShareGPT.com to train its models. Introducing the Evol-Instruct method, WizardLM~\cite{xu2023wizardlm} involves evolving existing instruction data to generate more intricate and diverse datasets. 

\textbf{Chain-of-Thought Instruction Tuning.} Contrary to general methods, recent research~\cite{wizardcoder, yue2023mammoth, chen2022program, gunasekar2023textbooks, haluptzok2023language} employs instruction tuning in various domains such as common-sense reasoning~\cite{west-etal-2022-symbolic}, text-summarization~\cite{sclar2022referee}, and mathematical reasoning~\cite{luo2023wizardmath,yue2023mammoth}. It's also applied in tool use~\cite{patil2023gorilla}, coding~\cite{wizardcoder}, and universal reasoning~\cite{li-etal-2023-symbolic, zelikman2022star}. 
Among them, ~\cite{yue2023mammoth} offers a diverse math problem corpus with annotations similar to our module-of-thought, using chain-of-thought or program-of-thought~\cite{chen2022program}. \cite{gunasekar2023textbooks} suggests pre-training models on artificially created programming textbooks from GPT3.5. In a similar vein, \cite{haluptzok2023language} generates coding puzzles and their solutions using language models.

\subsection{Prompting Techniques}
The Chain of Thought (CoT) technique \cite{wei2022chain} introduces an approach for language reasoning tasks by generating intermediate reasoning steps before providing the final answer. Subsequent approach least-to-most prompting \cite{leastmost}, simplifies a complex problem into a sequence of smaller sub-problems, solving them sequentially and incorporating the solution of each preceding sub-problem into the prompt for the next. Furthermore, PAL \cite{pal} and PoT \cite{POT} use code generation to create intermediate reasoning steps. 
Similar methods are proposed for simple mathematical \cite{LewkowyczADDMRS22,WuJLRSJS22}, commonsense \cite{T0,MadaanZ0YN22}, symbolic reasoning \cite{YaoZYDSN023} and code generation problems~\cite{self-planning, codechain}. However, these works can only plan code during generation. In comparison, our approach introduces a guided module-of-thought framework during training, making it more intrinsic. Our investigations reveal that, through the cultivation and utilization of sub-modules, MoTCoder significantly enhances both the modularity and correctness of the generated solutions.


\section{Methods}
\label{sec:methods}

In this section, we first detail the module-of-thought instruction transformation in \cref{sec:transformation} and then introduce the module-of-thought instruction tuning strategy for our MoTCoder in \cref{sec:tuning}.

\subsection{Module-of-Thought Instruction Transformation}
\label{sec:transformation}
In this section, we first introduce the normal instruction, followed by our module-of-thought instruction and its assessment.

\subsubsection{Normal Instruction}
In general, a code sequence generated by a language model $\theta$ through the autoregressive sampling of tokens $\hat{o}_t$ is from the parameterized conditional distribution:
\begin{equation}
    o_t \sim p_\theta (.| o_{1:t-1}, I),
\end{equation}
where $I$ represents the input instruction and $o_{t}$ is the $t$-th token of the flattened output sequence.

\subsubsection{Module-of-Thought Instruction} 
Considering programming encompasses both detailed coding abilities and broader logical or strategic planning capabilities, we propose a transformation pipeline that enhances code clarity and structure targeting detailed coding proficiencies and integrating descriptions based on natural language for the broader strategic planning capabilities. 

According to previous researches~\cite{jain2023llmassisted}, straightforward and unambiguous, detailed instructions enhance the model's efficacy and precision in executing the desired tasks. Hence, for intricate data transformation tasks, decomposing the task into simpler, sequential steps yields better outcomes. Therefore, our proposed methodology aims to transform normal instructions into a sequential code generation process by leading the models through a two-step procedure, as illustrated in \cref{fig:framework} .

\begin{enumerate}
    \item \textbf{Sub-modules.} Initially, the models are instructed to outline the required sub-modules, generating only their function headers and docstrings describing their intended usage.
    \begin{equation}
    \hat{S_i} \sim p_\theta (.| \hat{S}_{1:i-1}, I), \label{eq:sub}
    \end{equation}
    where $\hat{S}_i$ represents the $i$-th sub-module outlined by the model and $I$ represents the input instruction. 
    \item \textbf{Final solution.} The subsequent instruction guides the model to implement these sub-modules and eventually combine them into a comprehensive final solution. 
    \begin{equation}
    \hat{o}_t \sim p_\theta (.| \hat{o}_{1:t-1}, \{\hat{S}_i\}, I),
    \end{equation}
    where $\hat{o}_{t}$ is the $t$-th of the flattened output sequence. 
\end{enumerate}

The instruction is supplemented with a one-shot example, serving to prompt the model to adhere to the MoT instruction generation strategy. An illustration of the instruction prompt is presented in \cref{fig:prompts}. The instruction encourages the model to decompose a program into sub-modules. This mirrors the methodology commonly employed by developers when addressing intricate coding tasks, where they systematically break down solutions into modular components.

\subsubsection{Instruction Assessment} Throughout these transformations, we introduce guidelines for reviewing our transformed module-of-thought code. We identify the situations below as markers of instruction refinement failure:

\begin{enumerate}
\item The refined instruction diverges from the module-of-thought generation strategy, not adhering to the protocol of initial sub-module creation followed by the main code development.
   
\item At the sub-module creation phase, if no sub-modules are formed or if overarching code is developed instead.

\item During the main code development phase, the absence of main code creation or the emergence of multiple main code blocks indicates a problem.

\item The presence of test cases in the dataset that the transformed program fails to pass. This criterion ensures the transformed programs preserve functional equivalence with the original codes.

\end{enumerate}

\subsection{Module-of-Thought Instruction Tuning}
\label{sec:tuning}
In this section, we first introduce our module-of-thought dataset, followed by our MoT instruction tuning details.

\subsubsection{MoT Dataset} 

\re{Our queries are sourced from the training sets of APPS~\cite{hendrycksapps2021} and CodeContests~\cite{AlphaCode}. There is overlap between the CodeContests training set and the APPS test set, so we performed detailed deduplication to prevent test data leakage. For each problem, we take at most 100 answers. We then use GPT4o to generate transformed instructions. We include two types of data: clean and \idea (MoT). The goals for clean data are: 1. Optimize variable names to better reflect their purpose. 2. Add comments. 3. Follow the instructions and the meaning of the original code without changing its functionality. MOT data additionally uses functions if there are code segments that are functionally clear and reusable. All generated data are tested using input-output examples from the training set, and any data not passing the test are discarded. As a result, we have collected a total of 183K clean code data and 174K MoT data for our final training dataset. Detailed statistics of the data are shown in \cref{tab:data_statistic}.}

\begin{table}[t]
    \normalsize
    \centering
    \setlength{\tabcolsep}{1mm}
    \caption{\re{Data Statistics of our training dataset.}}
    \begin{tabular}{lcccc}
        \toprule
        Data & Data & Pre-filtering & Post-filtering & Passing  \\
        Type & Source & Count & Count & Rate \\
        \midrule
        \multirow{2}{*}{MoT} & APPS & 117K & 68K & 58\% \\
                             & CodeContests & 154K & 70K & 45\% \\
        \multirow{2}{*}{Clean} & APPS & 117K & 75K & 64\% \\
                               & CodeContests & 154K & 99K & 64\% \\
        \bottomrule
    \end{tabular}
    \label{tab:data_statistic}
\end{table}

\subsubsection{MoT Instruction Tuning} 
\re{Our MoTCoder underwent instruction tuning utilizing the SOTA coding model Qwen2.5-Coder-7B-Instruct~\cite{qwen25coder} across one epoch on our proposed MoT instruction dataset, and the best-performing model during the training process was selected. The model's maximum input length was 2048 tokens. We used a training batch size of 16 and an evaluation batch size of 4 per device, with gradient accumulation steps set to 4. The learning rate was initialized at \(2 \times 10^{-6}\), with a warmup ratio of 0.03 steps to gradually adapt the learning rate}, following a cosine learning rate scheduler for optimization. Additionally, our model was configured with the AdamW optimizer and utilized the WarmupLR scheduler to manage the learning rate adjustments effectively. To optimize memory and compute resources, we employed a third-stage zero optimization setting and enable communication overlap, contiguous gradients, and large sub group size of \(1 \times 10^9\) to streamline the training process. In addition, the maximum live parameters, maximum reuse distance, and the parameter settings for gathering 16-bit weights during model saving were all set to \(1 \times 10^9\).


\begin{table*}[t]
    \centering
    \normalsize
    \caption{\re{APPs test results by \emph{pass@1} (\%) of the ablation experiment on training datasets comparing MoT and normal finetuning.}}
    \label{tab:dataset}
    \begin{tabular}{lr|cccc}
        \toprule
        Model & Size & Introductory & Interview & Competition & All \\
        \midrule
        Qwen2.5-Coder-Instruct & 7B & 50.58 & 30.32 & 19.49 & 32.21 \\
        Normal Finetuning & 7B & 45.36 & 25.74 & 15.92 & 27.70 \\
        \rowcolor{teal!20}MoT Finetuning & 7B & \textbf{54.26} & \textbf{32.63} & \textbf{21.18} & \textbf{34.67} \\
        \bottomrule
    \end{tabular}
\end{table*}


\begin{table*}[t!]
 \center
 \normalsize
 \caption{\textbf{APPS test results by \emph{pass@k} (\%).} Competitive approaches include CodeT5~\cite{codet5}, fine-tuned GPT-Neo~\cite{hendrycks2021measuring}, GPT-2~\cite{gpt2}, GPT-3~\cite{GPT3}, one-shot StarCoder~\cite{starcoder}, WizardCoder~\cite{wizardcoder}, CodeLlama series~\cite{codellama}, text-davinci, \re{Deepseek-Coder~\cite{deepseekcoder} and Qwen2.5-coder~\cite{qwen25coder} and GPT4o~\cite{openai2023gpt4}}. We also include code-revision methods containing Self-edit~\cite{zhang2023self}, CodeRL~\cite{le2022coderl}, Self-repair~\cite{olausson2023demystifying}, and CodeChain~\cite{codechain}. }
 \setlength{\tabcolsep}{3mm}
 \begin{tabular}{lr|cccc} 
 \toprule
  Model & Size & Introductory & Interview & Competition & All \\
  \midrule
CodeT5 & 770M  & 6.60 & 1.03 & 0.30 & 2.00 \\
CodeRL+CodeT5 & 770M & 7.08 & 1.86 & 0.75 & 2.69 \\ 
text-davinci-002 & - & - & - & - & 7.48 \\
Self-edit+text-davinci-002 & - & - & - & - & 7.94 \\
GPT-2 & 0.1B & 5.64 & 6.93 & 4.37 & 6.16 \\
 & 1.5B & 7.40 & 9.11 & 5.05 & 7.96 \\ 
 GPT-Neo & 2.7B & 14.68 & 9.85 & 6.54 & 10.15\\ 
GPT-3 & 175B & 0.57 & 0.65 & 0.21 & 0.55 \\
StarCoder & 15B & 7.25 & 6.89 & 4.08 & 6.40 \\ 
WizardCoder & 15B & 26.04 & 4.21 & 0.81 & 7.90 \\
CodeChain+WizardCoder & 15B & 26.29 & 7.49 & 3.75 & 10.50 \\ 
Octocoder & 16B & 16.50 & 7.92 & 4.61 & 8.97 \\
CodeLlama & 7B & 14.15 & 6.63 & 4.00 & 7.61 \\
 & 13B & 23.94 & 13.50 & 9.80 & 14.85 \\ 
 & 34B & 32.01 & 18.61 & 10.19 & 19.61 \\ 
CodeLlama-Python & 7B & 18.83 & 8.62 & 4.47 & 9.83 \\
 & 13B & 26.40 & 13.44 & 6.86 & 14.72 \\
 & 34B & 26.45 & 16.61 & 8.77 & 17.01 \\ 
CodeLlama-Instruct & 7B & 14.20 & 6.63 & 4.43 & 7.70 \\
 & 13B & 22.41 & 14.34 & 6.62 & 15.21 \\
 & 34B & 28.64 & 16.80 & 10.51 & 17.91 \\ 
Deepseek-Coder-Base & 6.7B & 40.23 & 22.12 & 13.04 & 23.92 \\
Deepseek-Coder-Instruct & 6.7B & 44.65 & 23.86 & 12.89 & 25.83 \\
\re{Qwen2.5-Coder }          & \re{7B}   & \re{51.33}        & \re{29.37}     & \re{17.54}       & \re{31.40} \\
& \rv{32B} & \rv{61.00} & \rv{37.47} & \rv{21.50} & \rv{38.98} \\
\re{Qwen2.5-Coder-Instruct}  & \re{7B}     & \re{50.58}       & \re{30.32}     & \re{19.49}       & \re{32.21} \\        
& \rv{32B} & \rv{60.72} & \rv{38.60} & \rv{24.11} & \rv{40.13} \\
\rowcolor{teal!20}\re{MoTCoder}    & \re{7B}     & \re{54.26}       & \re{32.63}     & \re{21.18}       & \re{34.67} \\
\rowcolor{teal!20}& \rv{32B} & \rv{\textbf{68.44}} & \rv{\textbf{44.49}} & \rv{\textbf{27.84}} & \rv{\textbf{45.95}} \\
\re{GPT4o}                   & \re{-}     & \re{78.53}        & \re{55.57}     & \re{31.46}       & \re{55.34} \\
 \bottomrule
 \end{tabular}
 \label{tab:apps_test}
\end{table*}

\begin{table}[t]
    \centering
    \normalsize
    \setlength{\tabcolsep}{1.2mm}
    \caption{\re{CodeContests test and validation results by \emph{pass@k} (\%). Competing models include StarCoder~\cite{starcoder}, WizardCoder~\cite{wizardcoder}, CodeLlama series models~\cite{codellama}, Deepseek-Coder~\cite{deepseekcoder}, Qwen2.5-Coder~\cite{qwen25coder} and GPT4o~\cite{openai2023gpt4}.}}
    \begin{tabular}{lr|cccc}
        \toprule
        Model & Size & Test & Valid & All \\
        \midrule
        WizardCoder & 15B & 0.44 & 0.17 & 0.33 \\
        CodeLlama-Instruct & 7B & 5.57 & 2.26 & 4.20 \\
        & 13B & 5.52 & 3.80 & 4.81 \\
        & 34B & 5.57 & 3.87 & 4.86 \\
        CodeLlama-Python & 7B & 1.80 & 1.24 & 1.57 \\
        & 13B & 4.33 & 0.92 & 2.89 \\
        & 34B & 3.88 & 1.50 & 2.92 \\
        CodeLlama & 7B & 0.13 & 0.00 & 0.08 \\
        & 13B & 7.62 & 2.80 & 4.18\\
        & 34B & 5.13 & 2.85 &  5.62 \\
        Deepseek-Coder-Base & 6.7B & 11.48 & 5.78 & 9.12 \\
        Deepseek-Coder-Instruct & 6.7B & 15.25 & 10.10 & 13.11 \\
        Qwen2.5-Coder & 7B & 16.41 & 13.49 & 15.20 \\
         & \rv{32B} & \rv{20.41} & \rv{14.41} & \rv{17.92} \\
        Qwen2.5-Coder-Instruct & 7B & 15.45 & 13.42 & 14.61 \\
         & \rv{32B} & \rv{12.67} & \rv{14.40} & \rv{13.39} \\
        \rowcolor{teal!20}MoTCoder & 7B & 20.77 & 16.72 & 19.09 \\
        \rowcolor{teal!20} & \rv{32B} & \rv{\textbf{26.34}} & \rv{\textbf{20.35}} & \rv{\textbf{23.85}} \\
        GPT4o & - & 29.76 & 30.42 & 30.03 \\
        \bottomrule
    \end{tabular}
    \label{tab:codecontests}
\end{table}

\begin{table*}[t]
    \centering
    \normalsize
    \caption{\re{Model performance comparing MoTCoder and Qwen2.5-Coder-Instruct with self-reflection on CodeContests.}}
    \setlength{\tabcolsep}{3mm}
    \begin{tabular}{lr|ccc}
        \toprule
        Model & Size & Validation & Test & All \\
        \midrule
        Qwen2.5-Coder-Instruct & 7B & 13.42 & 15.45 & 14.61 \\
        Qwen2.5-Coder-Instruct + Self-Reflection & 7B & 19.88 & 19.74 & 19.80 \\
        \rowcolor{teal!20}MoTCoder & 7B & \textbf{16.72} & \textbf{20.77} & \textbf{19.09} \\
        \rowcolor{teal!20}MoTCoder + Self-Reflection & 7B & \textbf{21.63} & \textbf{24.10} & \textbf{23.08} \\
        \bottomrule
    \end{tabular}
    \label{tab:reflection}
\end{table*}

\section{Experiments}\label{sec:exp}

We demonstrate the efficacy of MoTCoder in tackling intricate code-generation tasks, rather than those that can be solved with just a few lines, as exemplified by HumanEval~\cite{humeval} and MBPP~\cite{MBPP}. Specifically, we focus on the following prominent benchmarks.

\textbf{APPS}~\cite{hendrycksapps2021} is a description-to-code generation benchmark from competitive programming platforms Codewars\footnote{\url{https://www.codewars.com/}}, AtCoder\footnote{\url{https://atcoder.jp/}}, Kattis\footnote{\url{https://open.kattis.com/}}, Codeforces\footnote{\url{https://codeforces.com/}}, etc. Building upon prior research~\cite{hendrycksapps2021, humeval, AlphaCode}, we conducted an assessment of the models using the passing rate metric \emph{pass@k}. This metric is defined as the proportion of problems successfully solved by employing \emph{k} generated programs for each problem. 
    
\textbf{CodeContests}~\cite{AlphaCode} is a competitive programming dataset sourced from Aizu\footnote{\url{https://judge.u-aizu.ac.jp}}, AtCoder\footnote{\url{https://atcoder.jp}}, CodeChef\footnote{\url{https://www.codechef.com}}, Codeforces, HackerEarth\footnote{\url{https://www.hackerearth.com}}, etc. Building upon prior research~\cite{hendrycksapps2021, humeval, AlphaCode}, we conducted an assessment of the models using the passing rate metric \emph{pass@k}.

\subsection{Ablation Experiments} 
\label{sec:ablation}

In this section, we conduct ablation experiments to investigate the effects of MoT and normal finetuning. \re{We use the Qwen2.5-Coder-7B-Instruct model as the base model. To control variables, we apply the same parameters and instructions during finetuning for both approaches. For ground truth, we use our constructed MoT code for the MoT finetuning, and standard code from the APPS and CodeContests training datasets for the normal finetuning.}

\re{The results of these experiments are presented in \cref{tab:dataset}. It is intriguing to note that, for a well-trained model like Qwen2.5-Coder-7B-Instruct, finetuning with uncurated normal data does not further enhance performance; rather, it leads to a decline, with scores dropping from 32.21 to 27.70 in terms of \emph{pass@1} accuracy. In contrast, when performed with our MoT finetune, the model exhibits significant improvements across all levels, as evidenced by the enhanced scores. Overall, the application of MoT finetuning results in an improvement from 32.21\% to 34.67\% in the overall \emph{pass@1} accuracy. This increase highlights the capacity of our MoT method to further augment the proficiency of a well-trained model. }


\subsection{Results on APPS} 
\label{sec:res_apps}

We conducted a comparison of our approach with existing large language model baselines on APPs~\cite{hendrycks2021measuring}. All outcomes are computed using raw predictions without being filtered by the test cases provided in the prompt. Our analysis includes a comparison with open-sourced approaches such as CodeT5~\cite{codet5}, fine-tuned GPT-Neo~\cite{hendrycks2021measuring}, GPT-2~\cite{gpt2}, GPT-3~\cite{GPT3}, one-shot StarCoder~\cite{starcoder}, WizardCoder~\cite{wizardcoder}, CodeLlama series~\cite{codellama}, \re{Deepseek-Coder~\cite{deepseekcoder}, and Qwen2.5-Coder~\cite{qwen25coder}. Additionally, we present results from SOTA closed-source model GPT4o~\cite{openai2023gpt4}.}

As depicted in the results from the APPS shown in \cref{tab:apps_test}, we notice that the module-of-thought training approach leads to an enhancement in code generation capabilities compared with previous instruction finetuning models. Our MoTCoder exhibits improved performance across all difficulty levels and demonstrates more substantial gains in the interview and competition-level problems, characterized by more intricate solutions. To provide specifics, the \emph{pass@1} performance of MoTCoder-6.7B  surprisingly outperformed the closed-source model GPT-4 by an impressive margin of 12.61\%. This outcome serves as compelling evidence of the efficacy of our approach in addressing competitive programming problems.

We also conduct a comparative analysis of our approach against previous LLM baselines with code-revision methods as well. Our included baselines contain Codex~\cite{humeval}, CodeT5~\cite{codet5}, code-davinci, StarCoder~\cite{starcoder}, and WizardCoder~\cite{wizardcoder} and code-revision methods contain Self-edit~\cite{zhang2023self}, CodeRL~\cite{codet5, le2022coderl}, Self-repair~\cite{olausson2023demystifying}, and CodeChain~\cite{codechain}. The results presented in \cref{tab:apps_test} illustrate that MoTCoder exhibits notable performance improvements. 

\rv{MoTCoder demonstrates superior performance across all categories compared to Qwen2.5-Coder-Instruct, achieving higher scores on the all of difficulty levels. MoTCoder-7B achieves an average score of 34.67\%, surpassing Qwen2.5-Coder-Instruct-7B by 2.46\%. MoTCoder-32B reaches an average score of 45.95\%, surpassing the baseline model by 5.82\%. This improvement highlights the effectiveness of MoTCoder's guided module-of-thought framework.}

\subsection{Results on CodeContests} 
\label{sec:res_cc}

We conduct an evaluation of our approach on CodeContests~\cite{AlphaCode}, benchmarking it against current coding models, including StarCoder~\cite{starcoder}, WizardCoder~\cite{wizardcoder}, CodeLlama series models~\cite{codellama}, \re{Deepseek-Coder~\cite{deepseekcoder}, Qwen2.5-Coder~\cite{qwen25coder}. Furthermore, we present results from SOTA closed-source model GPT4o~\cite{openai2023gpt4}. The results, as depicted in \cref{tab:codecontests}, reveal notable performance enhancements achieved by MoTCoder. }
\rv{Specifically, MoTCoder-7B achieves the performance of 19.09\% compared to Qwen2.5-Coder (+3.89\%), and MoTCoder-32B reaches 23.85\% (+5.93\%). These results demonstrate MoTCoder's superior capability in generating accurate code solutions.}

\subsection{Results on Self-Reflection}

\re{To further explore our model's interactive and self-corrective capabilities, we constructed a multi-turn dialogue task. For cases in the CodeContests~\cite{AlphaCode} where the model did not succeed in passing all test cases, we prompted the model to self-reflect and regenerate the solutions. There is up to 5 rounds of reflection.}

\re{The results in \cref{tab:reflection} demonstrate that our models achieved significant improvements in both code accuracy and self-correction capabilities. Specifically, the Qwen2.5-Coder-Instruct model with self-reflection showed an increase from 14.61\% to 19.80\% in the overall score. Moreover, the MoTCoder-7B model achieved comparable performance to Qwen2.5-Coder-Instruct without the need for reflection. With self-reflection, MoTCoder-7B improved from 19.09\% to 23.08\%, surpassing Qwen2.5-Coder-Instruct by 3.28\%.}


\section{Further Analysis}
\label{sec:discuss}
\re{In this section, we will first analyze the relationship between problem complexity and optimal module decomposition (\cref{sec:acc_func}). Then, we will conduct a quantitative analysis of the time and memory consumption, verifying that our approach can reduce memory consumption (\cref{sec:time_mem}). Finally, we will use the maintainability index to validate structural quality of the generated solutions(\cref{sec:reusage}).}

\subsection{Influence of modules on Accuracy}
\label{sec:acc_func}

\begin{figure*}[t]
    \centering
    \begin{subfigure}[t]{0.33\textwidth}
        \centering
        \includegraphics[width=\linewidth]{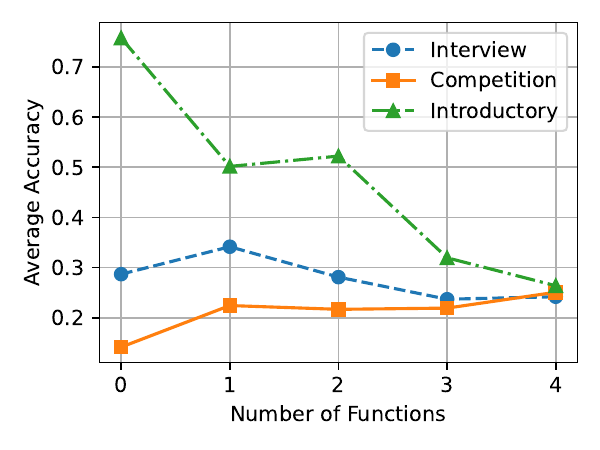} 
        \caption{}
    \end{subfigure}%
    \hfill
    \begin{subfigure}[t]{0.33\textwidth}
        \centering
        \includegraphics[width=\linewidth]{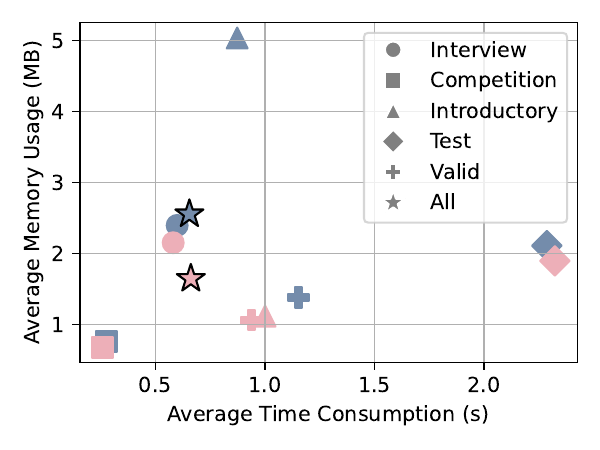}
        \caption{}
    \end{subfigure}%
    \hfill
    \begin{subfigure}[t]{0.33\textwidth}
        \centering
\includegraphics[width=\linewidth]{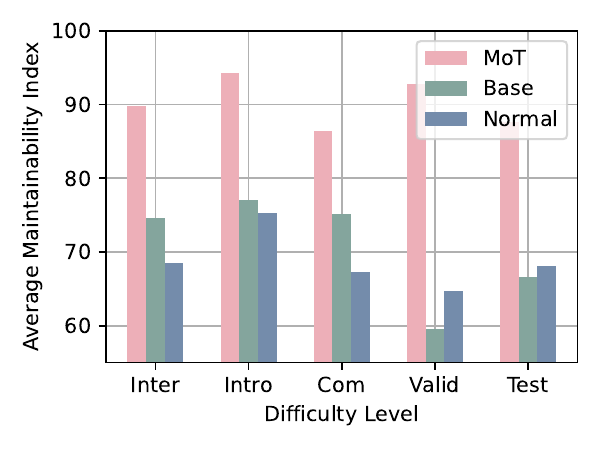}
        \caption{}
    \end{subfigure}
    \caption{\re{(a) Relationship between the number of functions and accuracy for solutions generated by MoTCoder across different difficulty levels in the APPS dataset. (b) Average time and memory consumption for the passed output for MoT (pink) and Normal (blue). (c) Average maintainability index for MoT finetuning, normal finetuning and baseline models. Valid and test are from CodeContests. Interview, introductory and competition are from APPS test set.}}
    \label{fig:combined_figures}
\end{figure*}

\re{We conducted an analysis using the code generated by MoTCoder to evaluate the number of functions in the solution codes and their impact on accuracy across different difficulty levels. We focus on the relationship between problem complexity and optimal modular decomposition. The difficulty levels are categorized as introductory, interview, and competition, moving from easiest to hardest, as depicted in \cref{fig:combined_figures}(a).}

\re{The results reveal distinct trends across difficulty levels. For introductory problems, the accuracy generally declines as the number of functions increases, suggesting that simpler solutions benefit from minimal modularity. In contrast, the interview level maintains relatively stable accuracy across different function counts. Notably, for competition-level problems, there is an upward trend in accuracy with an increase in the number of functions.}

\re{This observation indicates that while simpler problems are best addressed using fewer, singular modules, more complex problems benefit from being broken down into modular functions. This aligns with our intuition: for straightforward problems that can be solved with just a few lines of code, excessive modularization adds unnecessary complexity. Conversely, for challenging problems, decomposing them into submodules facilitates a more effective solution process, reflecting human strategies for tackling difficult issues.}

\subsection{Time and Memory Consumption}
\label{sec:time_mem}
\re{In \cref{fig:combined_figures}(b), we analyze the average time and memory consumption for the generated MoT code and normal code, which are produced by the MoT finetuning and normal finetuning models, respectively. We collected data on samples from APPS and CodeContests where both MoT and normal code passed. It is evident that while the time consumption for the MoT model is comparable to that of the normal model, MoT finetuning consistently shows significantly lower memory consumption at all levels. This indicates that MoT finetuning is more memory efficient, as it is able to efficiently release unused memory by distinguishing between global variables and local variables that are only used within functions. Therefore, it is advantageous for scenarios with memory constraints.}

\subsection{Maintainability Analysis }
\label{sec:reusage}

\re{In this section, we explore the maintainability metrics of the models. We compare the models after MoT finetuning and normal finetuning, as well as the baseline model Qwen2.5-Coder-7B-Instruct. We collected statistics on the generated passing code of these models on APPS and CodeContest, using the radon~\cite{radon} tool to calculate their maintainability index. }

\re{The maintainability index is a composite metric that evaluates the ease with which code can be maintained. It is calculated using the following formula:}

\begin{equation}
\begin{aligned}
MI &= \max(0, 171 - 5.2 \times \log_2(V) - 0.23 \times CC \\ 
&- 16.2 \times \log_2(LOC) 
+ 50 \times \sin(\sqrt{2.46 \times C}))
\end{aligned}
\end{equation}

\re{where}:
\begin{itemize}
    \item \re{$V$ is the Halstead Volume, representing the size of the program.}
    \item \re{$CC$ is the Cyclomatic Complexity, indicating the number of independent paths through the code.}
    \item \re{$LOC$ is the Lines of Code, reflecting the physical line count in the source code.}
    \item \re{$C$ is the Comment Density, the ratio of comment lines to total lines.}
\end{itemize}

\re{High maintainability index values generally indicate that the code is easier to maintain, usually due to lower complexity, fewer lines of code, and adequate documentation.}

\re{The results, displayed in \cref{fig:combined_figures}(c), show that for all levels, MoT code consistently demonstrates significantly higher maintainability compared to normal finetuning code and the baseline. This suggests that MoT finetuning leads to code that is easier to understand and modify, which can be particularly beneficial in scenarios requiring long-term code maintenance and evolution.}

\section{Conclusion}\label{sec:conclusion}
This study highlights the limitations of Large Language Models (LLMs) in solving complex programming tasks due to their tendency to generate monolithic code blocks. In response, we developed Module-of-Thought Coder (MoTCoder), a framework that encourages the breakdown of tasks into manageable sub-tasks and sub-modules. Our results demonstrate that MoTCoder's approach significantly enhances the modularity and accuracy of solutions, as evidenced by considerable improvements in \emph{pass@1} rates on both APPS and CodeContests benchmarks. \re{Through the process of MoT instruction tuning, MoTCoder also achieved notable advancements in self-correction capabilities. Additionally, our analysis shows that MoTCoder improves the maintainability index of the generated code, thereby making it easier to comprehend and modify.} We believe the introduction of MoT instruction tuning as a method to cultivate and leverage sub-modules paves the path for a promising direction for future research. 

\bibliographystyle{IEEEtran}
\bibliography{egbib}

\begin{thebibliography}{10}
\providecommand{\url}[1]{#1}
\csname url@samestyle\endcsname
\providecommand{\newblock}{\relax}
\providecommand{\bibinfo}[2]{#2}
\providecommand{\BIBentrySTDinterwordspacing}{\spaceskip=0pt\relax}
\providecommand{\BIBentryALTinterwordstretchfactor}{4}
\providecommand{\BIBentryALTinterwordspacing}{\spaceskip=\fontdimen2\font plus
\BIBentryALTinterwordstretchfactor\fontdimen3\font minus \fontdimen4\font\relax}
\providecommand{\BIBforeignlanguage}[2]{{%
\expandafter\ifx\csname l@#1\endcsname\relax
\typeout{** WARNING: IEEEtran.bst: No hyphenation pattern has been}%
\typeout{** loaded for the language `#1'. Using the pattern for}%
\typeout{** the default language instead.}%
\else
\language=\csname l@#1\endcsname
\fi
#2}}
\providecommand{\BIBdecl}{\relax}
\BIBdecl

\bibitem{Zohar71}
\BIBentryALTinterwordspacing
Z.~Manna and R.~J. Waldinger, ``Toward automatic program synthesis,'' \emph{Commun. ACM}, vol.~14, no.~3, p. 151–165, mar 1971. [Online]. Available: \url{https://doi.org/10.1145/362566.362568}
\BIBentrySTDinterwordspacing

\bibitem{GPT3}
T.~B. Brown, B.~Mann, N.~Ryder, M.~Subbiah, J.~Kaplan, P.~Dhariwal, A.~Neelakantan, P.~Shyam, G.~Sastry, A.~Askell, S.~Agarwal, A.~Herbert{-}Voss, G.~Krueger, T.~Henighan, R.~Child, A.~Ramesh, D.~M. Ziegler, J.~Wu, C.~Winter, C.~Hesse, M.~Chen, E.~Sigler, M.~Litwin, S.~Gray, B.~Chess, J.~Clark, C.~Berner, S.~McCandlish, A.~Radford, I.~Sutskever, and D.~Amodei, ``Language models are few-shot learners,'' in \emph{Advances in Neural Information Processing Systems 33: Annual Conference on Neural Information Processing Systems 2020, NeurIPS 2020, December 6-12, 2020, virtual}, H.~Larochelle, M.~Ranzato, R.~Hadsell, M.~Balcan, and H.~Lin, Eds., 2020.

\bibitem{GPT4}
\BIBentryALTinterwordspacing
OpenAI, ``{GPT-4} technical report,'' \emph{CoRR}, vol. abs/2303.08774, 2023. [Online]. Available: \url{https://doi.org/10.48550/arXiv.2303.08774}
\BIBentrySTDinterwordspacing

\bibitem{PaLM}
A.~Chowdhery, S.~Narang, J.~Devlin, M.~Bosma, G.~Mishra, A.~Roberts, P.~Barham, H.~W. Chung, C.~Sutton, S.~Gehrmann, P.~Schuh, K.~Shi, S.~Tsvyashchenko, J.~Maynez, A.~Rao, P.~Barnes, Y.~Tay, N.~Shazeer, V.~Prabhakaran, E.~Reif, N.~Du, B.~Hutchinson, R.~Pope, J.~Bradbury, J.~Austin, M.~Isard, G.~Gur{-}Ari, P.~Yin, T.~Duke, A.~Levskaya, S.~Ghemawat, S.~Dev, H.~Michalewski, X.~Garcia, V.~Misra, K.~Robinson, L.~Fedus, D.~Zhou, D.~Ippolito, D.~Luan, H.~Lim, B.~Zoph, A.~Spiridonov, R.~Sepassi, D.~Dohan, S.~Agrawal, M.~Omernick, A.~M. Dai, T.~S. Pillai, M.~Pellat, A.~Lewkowycz, E.~Moreira, R.~Child, O.~Polozov, K.~Lee, Z.~Zhou, X.~Wang, B.~Saeta, M.~Diaz, O.~Firat, M.~Catasta, J.~Wei, K.~Meier{-}Hellstern, D.~Eck, J.~Dean, S.~Petrov, and N.~Fiedel, ``Palm: Scaling language modeling with pathways,'' \emph{CoRR}, vol. abs/2204.02311, 2022.

\bibitem{palm2}
\BIBentryALTinterwordspacing
R.~Anil, A.~M. Dai, O.~Firat, M.~Johnson, D.~Lepikhin, A.~Passos, S.~Shakeri, E.~Taropa, P.~Bailey, Z.~Chen, E.~Chu, J.~H. Clark, L.~E. Shafey, Y.~Huang, K.~Meier{-}Hellstern, G.~Mishra, E.~Moreira, M.~Omernick, K.~Robinson, S.~Ruder, Y.~Tay, K.~Xiao, Y.~Xu, Y.~Zhang, G.~H. {\'{A}}brego, J.~Ahn, J.~Austin, P.~Barham, J.~A. Botha, J.~Bradbury, S.~Brahma, K.~Brooks, M.~Catasta, Y.~Cheng, C.~Cherry, C.~A. Choquette{-}Choo, A.~Chowdhery, C.~Crepy, S.~Dave, M.~Dehghani, S.~Dev, J.~Devlin, M.~D{\'{\i}}az, N.~Du, E.~Dyer, V.~Feinberg, F.~Feng, V.~Fienber, M.~Freitag, X.~Garcia, S.~Gehrmann, L.~Gonzalez, and et~al., ``Palm 2 technical report,'' \emph{CoRR}, vol. abs/2305.10403, 2023. [Online]. Available: \url{https://doi.org/10.48550/arXiv.2305.10403}
\BIBentrySTDinterwordspacing

\bibitem{Chinchilla}
\BIBentryALTinterwordspacing
J.~Hoffmann, S.~Borgeaud, A.~Mensch, E.~Buchatskaya, T.~Cai, E.~Rutherford, D.~de~Las~Casas, L.~A. Hendricks, J.~Welbl, A.~Clark, T.~Hennigan, E.~Noland, K.~Millican, G.~van~den Driessche, B.~Damoc, A.~Guy, S.~Osindero, K.~Simonyan, E.~Elsen, J.~W. Rae, O.~Vinyals, and L.~Sifre, ``Training compute-optimal large language models,'' \emph{CoRR}, vol. abs/2203.15556, 2022. [Online]. Available: \url{https://doi.org/10.48550/arXiv.2203.15556}
\BIBentrySTDinterwordspacing

\bibitem{gopher}
\BIBentryALTinterwordspacing
J.~W. Rae, S.~Borgeaud, T.~Cai, K.~Millican, J.~Hoffmann, H.~F. Song, J.~Aslanides, S.~Henderson, R.~Ring, S.~Young, E.~Rutherford, T.~Hennigan, J.~Menick, A.~Cassirer, R.~Powell, G.~van~den Driessche, L.~A. Hendricks, M.~Rauh, P.~Huang, A.~Glaese, J.~Welbl, S.~Dathathri, S.~Huang, J.~Uesato, J.~Mellor, I.~Higgins, A.~Creswell, N.~McAleese, A.~Wu, E.~Elsen, S.~M. Jayakumar, E.~Buchatskaya, D.~Budden, E.~Sutherland, K.~Simonyan, M.~Paganini, L.~Sifre, L.~Martens, X.~L. Li, A.~Kuncoro, A.~Nematzadeh, E.~Gribovskaya, D.~Donato, A.~Lazaridou, A.~Mensch, J.~Lespiau, M.~Tsimpoukelli, N.~Grigorev, D.~Fritz, T.~Sottiaux, M.~Pajarskas, T.~Pohlen, Z.~Gong, D.~Toyama, C.~de~Masson~d'Autume, Y.~Li, T.~Terzi, V.~Mikulik, I.~Babuschkin, A.~Clark, D.~de~Las~Casas, A.~Guy, C.~Jones, J.~Bradbury, M.~J. Johnson, B.~A. Hechtman, L.~Weidinger, I.~Gabriel, W.~Isaac, E.~Lockhart, S.~Osindero, L.~Rimell, C.~Dyer, O.~Vinyals, K.~Ayoub, J.~Stanway, L.~Bennett, D.~Hassabis, K.~Kavukcuoglu, and G.~Irving, ``Scaling language models:
  Methods, analysis {\&} insights from training gopher,'' \emph{CoRR}, vol. abs/2112.11446, 2021. [Online]. Available: \url{https://arxiv.org/abs/2112.11446}
\BIBentrySTDinterwordspacing

\bibitem{GLM-130B}
\BIBentryALTinterwordspacing
A.~Zeng, X.~Liu, Z.~Du, Z.~Wang, H.~Lai, M.~Ding, Z.~Yang, Y.~Xu, W.~Zheng, X.~Xia, W.~L. Tam, Z.~Ma, Y.~Xue, J.~Zhai, W.~Chen, P.~Zhang, Y.~Dong, and J.~Tang, ``{GLM-130B:} an open bilingual pre-trained model,'' \emph{CoRR}, vol. abs/2210.02414, 2022. [Online]. Available: \url{https://doi.org/10.48550/arXiv.2210.02414}
\BIBentrySTDinterwordspacing

\bibitem{llama}
\BIBentryALTinterwordspacing
H.~Touvron, T.~Lavril, G.~Izacard, X.~Martinet, M.~Lachaux, T.~Lacroix, B.~Rozi{\`{e}}re, N.~Goyal, E.~Hambro, F.~Azhar, A.~Rodriguez, A.~Joulin, E.~Grave, and G.~Lample, ``Llama: Open and efficient foundation language models,'' \emph{CoRR}, vol. abs/2302.13971, 2023. [Online]. Available: \url{https://doi.org/10.48550/arXiv.2302.13971}
\BIBentrySTDinterwordspacing

\bibitem{opt}
\BIBentryALTinterwordspacing
S.~Zhang, S.~Roller, N.~Goyal, M.~Artetxe, M.~Chen, S.~Chen, C.~Dewan, M.~T. Diab, X.~Li, X.~V. Lin, T.~Mihaylov, M.~Ott, S.~Shleifer, K.~Shuster, D.~Simig, P.~S. Koura, A.~Sridhar, T.~Wang, and L.~Zettlemoyer, ``{OPT:} open pre-trained transformer language models,'' \emph{CoRR}, vol. abs/2205.01068, 2022. [Online]. Available: \url{https://doi.org/10.48550/arXiv.2205.01068}
\BIBentrySTDinterwordspacing

\bibitem{codellama}
B.~Roziere, J.~Gehring, F.~Gloeckle, S.~Sootla, I.~Gat, X.~E. Tan, Y.~Adi, J.~Liu, T.~Remez, J.~Rapin \emph{et~al.}, ``Code llama: Open foundation models for code,'' \emph{arXiv preprint arXiv:2308.12950}, 2023.

\bibitem{black10gpt}
S.~Black, G.~Leo, P.~Wang, C.~Leahy, and S.~Biderman, ``Gpt-neo: Large scale autoregressive language modeling with mesh-tensorflow,'' \emph{URL https://doi. org/10.5281/zenodo}, vol. 5297715, 2021.

\bibitem{humeval}
\BIBentryALTinterwordspacing
M.~Chen, J.~Tworek, H.~Jun, Q.~Yuan, H.~P. de~Oliveira~Pinto, J.~Kaplan, H.~Edwards, Y.~Burda, N.~Joseph, G.~Brockman, A.~Ray, R.~Puri, G.~Krueger, M.~Petrov, H.~Khlaaf, G.~Sastry, P.~Mishkin, B.~Chan, S.~Gray, N.~Ryder, M.~Pavlov, A.~Power, L.~Kaiser, M.~Bavarian, C.~Winter, P.~Tillet, F.~P. Such, D.~Cummings, M.~Plappert, F.~Chantzis, E.~Barnes, A.~Herbert{-}Voss, W.~H. Guss, A.~Nichol, A.~Paino, N.~Tezak, J.~Tang, I.~Babuschkin, S.~Balaji, S.~Jain, W.~Saunders, C.~Hesse, A.~N. Carr, J.~Leike, J.~Achiam, V.~Misra, E.~Morikawa, A.~Radford, M.~Knight, M.~Brundage, M.~Murati, K.~Mayer, P.~Welinder, B.~McGrew, D.~Amodei, S.~McCandlish, I.~Sutskever, and W.~Zaremba, ``Evaluating large language models trained on code,'' \emph{CoRR}, vol. abs/2107.03374, 2021. [Online]. Available: \url{https://arxiv.org/abs/2107.03374}
\BIBentrySTDinterwordspacing

\bibitem{starcoder}
R.~Li, L.~B. Allal, Y.~Zi, N.~Muennighoff, D.~Kocetkov, C.~Mou, M.~Marone, C.~Akiki, J.~Li, J.~Chim \emph{et~al.}, ``Starcoder: may the source be with you!'' \emph{arXiv preprint arXiv:2305.06161}, 2023.

\bibitem{wizardcoder}
Z.~Luo, C.~Xu, P.~Zhao, Q.~Sun, X.~Geng, W.~Hu, C.~Tao, J.~Ma, Q.~Lin, and D.~Jiang, ``Wizardcoder: Empowering code large language models with evol-instruct,'' 2023.

\bibitem{wang2023codet5+}
Y.~Wang, H.~Le, A.~D. Gotmare, N.~D. Bui, J.~Li, and S.~C. Hoi, ``Codet5+: Open code large language models for code understanding and generation,'' \emph{arXiv preprint arXiv:2305.07922}, 2023.

\bibitem{AlphaCode}
\BIBentryALTinterwordspacing
Y.~Li, D.~H. Choi, J.~Chung, N.~Kushman, J.~Schrittwieser, R.~Leblond, T.~Eccles, J.~Keeling, F.~Gimeno, A.~D. Lago, T.~Hubert, P.~Choy, C.~de~Masson~d'Autume, I.~Babuschkin, X.~Chen, P.~Huang, J.~Welbl, S.~Gowal, A.~Cherepanov, J.~Molloy, D.~J. Mankowitz, E.~S. Robson, P.~Kohli, N.~de~Freitas, K.~Kavukcuoglu, and O.~Vinyals, ``Competition-level code generation with alphacode,'' \emph{CoRR}, vol. abs/2203.07814, 2022. [Online]. Available: \url{https://doi.org/10.48550/arXiv.2203.07814}
\BIBentrySTDinterwordspacing

\bibitem{codegen}
E.~Nijkamp, B.~Pang, H.~Hayashi, L.~Tu, H.~Wang, Y.~Zhou, S.~Savarese, and C.~Xiong, ``Codegen: An open large language model for code with multi-turn program synthesis,'' in \emph{The Eleventh International Conference on Learning Representations}, 2023.

\bibitem{CodeGeeX}
\BIBentryALTinterwordspacing
Q.~Zheng, X.~Xia, X.~Zou, Y.~Dong, S.~Wang, Y.~Xue, Z.~Wang, L.~Shen, A.~Wang, Y.~Li, T.~Su, Z.~Yang, and J.~Tang, ``Codegeex: {A} pre-trained model for code generation with multilingual evaluations on humaneval-x,'' \emph{CoRR}, vol. abs/2303.17568, 2023. [Online]. Available: \url{https://doi.org/10.48550/arXiv.2303.17568}
\BIBentrySTDinterwordspacing

\bibitem{incoder}
\BIBentryALTinterwordspacing
D.~Fried, A.~Aghajanyan, J.~Lin, S.~Wang, E.~Wallace, F.~Shi, R.~Zhong, W.~Yih, L.~Zettlemoyer, and M.~Lewis, ``Incoder: {A} generative model for code infilling and synthesis,'' \emph{CoRR}, vol. abs/2204.05999, 2022. [Online]. Available: \url{https://doi.org/10.48550/arXiv.2204.05999}
\BIBentrySTDinterwordspacing

\bibitem{codet5}
Y.~Wang, W.~Wang, S.~R. Joty, and S.~C.~H. Hoi, ``Codet5: Identifier-aware unified pre-trained encoder-decoder models for code understanding and generation,'' in \emph{{EMNLP} {(1)}}.\hskip 1em plus 0.5em minus 0.4em\relax Association for Computational Linguistics, 2021, pp. 8696--8708.

\bibitem{MBPP}
\BIBentryALTinterwordspacing
J.~Austin, A.~Odena, M.~I. Nye, M.~Bosma, H.~Michalewski, D.~Dohan, E.~Jiang, C.~J. Cai, M.~Terry, Q.~V. Le, and C.~Sutton, ``Program synthesis with large language models,'' \emph{CoRR}, vol. abs/2108.07732, 2021. [Online]. Available: \url{https://arxiv.org/abs/2108.07732}
\BIBentrySTDinterwordspacing

\bibitem{hendrycksapps2021}
D.~Hendrycks, S.~Basart, S.~Kadavath, M.~Mazeika, A.~Arora, E.~Guo, C.~Burns, S.~Puranik, H.~He, D.~Song, and J.~Steinhardt, ``Measuring coding challenge competence with apps,'' \emph{NeurIPS}, 2021.

\bibitem{shinn2023reflexion}
N.~Shinn, F.~Cassano, B.~Labash, A.~Gopinath, K.~Narasimhan, and S.~Yao, ``Reflexion: Language agents with verbal reinforcement learning,'' 2023.

\bibitem{jiang2023selfplanning}
X.~Jiang, Y.~Dong, L.~Wang, Z.~Fang, Q.~Shang, G.~Li, Z.~Jin, and W.~Jiao, ``Self-planning code generation with large language models,'' 2023.

\bibitem{codechain}
H.~Le, H.~Chen, A.~Saha, A.~Gokul, D.~Sahoo, and S.~Joty, ``Codechain: Towards modular code generation through chain of self-revisions with representative sub-modules,'' \emph{arXiv preprint arXiv:2310.08992}, 2023.

\bibitem{GPT-NeoX-20B}
\BIBentryALTinterwordspacing
S.~Black, S.~Biderman, E.~Hallahan, Q.~Anthony, L.~Gao, L.~Golding, H.~He, C.~Leahy, K.~McDonell, J.~Phang, M.~Pieler, U.~S. Prashanth, S.~Purohit, L.~Reynolds, J.~Tow, B.~Wang, and S.~Weinbach, ``Gpt-neox-20b: An open-source autoregressive language model,'' \emph{CoRR}, vol. abs/2204.06745, 2022. [Online]. Available: \url{https://doi.org/10.48550/arXiv.2204.06745}
\BIBentrySTDinterwordspacing

\bibitem{gpt-j}
B.~Wang and A.~Komatsuzaki, ``{GPT-J-6B: A 6 Billion Parameter Autoregressive Language Model},'' \url{https://github.com/kingoflolz/mesh-transformer-jax}, May 2021.

\bibitem{UL2}
\BIBentryALTinterwordspacing
Y.~Tay, M.~Dehghani, V.~Q. Tran, X.~Garcia, D.~Bahri, T.~Schuster, H.~S. Zheng, N.~Houlsby, and D.~Metzler, ``Unifying language learning paradigms,'' \emph{CoRR}, vol. abs/2205.05131, 2022. [Online]. Available: \url{https://doi.org/10.48550/arXiv.2205.05131}
\BIBentrySTDinterwordspacing

\bibitem{qwen}
\BIBentryALTinterwordspacing
J.~Bai, S.~Bai, Y.~Chu, Z.~Cui, K.~Dang, X.~Deng, Y.~Fan, W.~Ge, Y.~Han, F.~Huang, B.~Hui, L.~Ji, M.~Li, J.~Lin, R.~Lin, D.~Liu, G.~Liu, C.~Lu, K.~Lu, J.~Ma, R.~Men, X.~Ren, X.~Ren, C.~Tan, S.~Tan, J.~Tu, P.~Wang, S.~Wang, W.~Wang, S.~Wu, B.~Xu, J.~Xu, A.~Yang, H.~Yang, J.~Yang, S.~Yang, Y.~Yao, B.~Yu, H.~Yuan, Z.~Yuan, J.~Zhang, X.~Zhang, Y.~Zhang, Z.~Zhang, C.~Zhou, J.~Zhou, X.~Zhou, and T.~Zhu, ``Qwen technical report,'' 2023. [Online]. Available: \url{https://arxiv.org/abs/2309.16609}
\BIBentrySTDinterwordspacing

\bibitem{qwen2}
\BIBentryALTinterwordspacing
A.~Yang, B.~Yang, B.~Hui, B.~Zheng, B.~Yu, C.~Zhou, C.~Li, C.~Li, D.~Liu, F.~Huang, G.~Dong, H.~Wei, H.~Lin, J.~Tang, J.~Wang, J.~Yang, J.~Tu, J.~Zhang, J.~Ma, J.~Yang, J.~Xu, J.~Zhou, J.~Bai, J.~He, J.~Lin, K.~Dang, K.~Lu, K.~Chen, K.~Yang, M.~Li, M.~Xue, N.~Ni, P.~Zhang, P.~Wang, R.~Peng, R.~Men, R.~Gao, R.~Lin, S.~Wang, S.~Bai, S.~Tan, T.~Zhu, T.~Li, T.~Liu, W.~Ge, X.~Deng, X.~Zhou, X.~Ren, X.~Zhang, X.~Wei, X.~Ren, X.~Liu, Y.~Fan, Y.~Yao, Y.~Zhang, Y.~Wan, Y.~Chu, Y.~Liu, Z.~Cui, Z.~Zhang, Z.~Guo, and Z.~Fan, ``Qwen2 technical report,'' 2024. [Online]. Available: \url{https://arxiv.org/abs/2407.10671}
\BIBentrySTDinterwordspacing

\bibitem{qwen25}
\BIBentryALTinterwordspacing
Qwen, :, A.~Yang, B.~Yang, B.~Zhang, B.~Hui, B.~Zheng, B.~Yu, C.~Li, D.~Liu, F.~Huang, H.~Wei, H.~Lin, J.~Yang, J.~Tu, J.~Zhang, J.~Yang, J.~Yang, J.~Zhou, J.~Lin, K.~Dang, K.~Lu, K.~Bao, K.~Yang, L.~Yu, M.~Li, M.~Xue, P.~Zhang, Q.~Zhu, R.~Men, R.~Lin, T.~Li, T.~Tang, T.~Xia, X.~Ren, X.~Ren, Y.~Fan, Y.~Su, Y.~Zhang, Y.~Wan, Y.~Liu, Z.~Cui, Z.~Zhang, and Z.~Qiu, ``Qwen2.5 technical report,'' 2025. [Online]. Available: \url{https://arxiv.org/abs/2412.15115}
\BIBentrySTDinterwordspacing

\bibitem{deepseek}
\BIBentryALTinterwordspacing
DeepSeek-AI, :, X.~Bi, D.~Chen, G.~Chen, S.~Chen, D.~Dai, C.~Deng, H.~Ding, K.~Dong, Q.~Du, Z.~Fu, H.~Gao, K.~Gao, W.~Gao, R.~Ge, K.~Guan, D.~Guo, J.~Guo, G.~Hao, Z.~Hao, Y.~He, W.~Hu, P.~Huang, E.~Li, G.~Li, J.~Li, Y.~Li, Y.~K. Li, W.~Liang, F.~Lin, A.~X. Liu, B.~Liu, W.~Liu, X.~Liu, X.~Liu, Y.~Liu, H.~Lu, S.~Lu, F.~Luo, S.~Ma, X.~Nie, T.~Pei, Y.~Piao, J.~Qiu, H.~Qu, T.~Ren, Z.~Ren, C.~Ruan, Z.~Sha, Z.~Shao, J.~Song, X.~Su, J.~Sun, Y.~Sun, M.~Tang, B.~Wang, P.~Wang, S.~Wang, Y.~Wang, Y.~Wang, T.~Wu, Y.~Wu, X.~Xie, Z.~Xie, Z.~Xie, Y.~Xiong, H.~Xu, R.~X. Xu, Y.~Xu, D.~Yang, Y.~You, S.~Yu, X.~Yu, B.~Zhang, H.~Zhang, L.~Zhang, L.~Zhang, M.~Zhang, M.~Zhang, W.~Zhang, Y.~Zhang, C.~Zhao, Y.~Zhao, S.~Zhou, S.~Zhou, Q.~Zhu, and Y.~Zou, ``Deepseek llm: Scaling open-source language models with longtermism,'' 2024. [Online]. Available: \url{https://arxiv.org/abs/2401.02954}
\BIBentrySTDinterwordspacing

\bibitem{deepseekv2}
\BIBentryALTinterwordspacing
DeepSeek-AI, A.~Liu, B.~Feng, B.~Wang, B.~Wang, B.~Liu, C.~Zhao, C.~Dengr, C.~Ruan, D.~Dai, D.~Guo, D.~Yang, D.~Chen, D.~Ji, E.~Li, F.~Lin, F.~Luo, G.~Hao, G.~Chen, G.~Li, H.~Zhang, H.~Xu, H.~Yang, H.~Zhang, H.~Ding, H.~Xin, H.~Gao, H.~Li, H.~Qu, J.~L. Cai, J.~Liang, J.~Guo, J.~Ni, J.~Li, J.~Chen, J.~Yuan, J.~Qiu, J.~Song, K.~Dong, K.~Gao, K.~Guan, L.~Wang, L.~Zhang, L.~Xu, L.~Xia, L.~Zhao, L.~Zhang, M.~Li, M.~Wang, M.~Zhang, M.~Zhang, M.~Tang, M.~Li, N.~Tian, P.~Huang, P.~Wang, P.~Zhang, Q.~Zhu, Q.~Chen, Q.~Du, R.~J. Chen, R.~L. Jin, R.~Ge, R.~Pan, R.~Xu, R.~Chen, S.~S. Li, S.~Lu, S.~Zhou, S.~Chen, S.~Wu, S.~Ye, S.~Ma, S.~Wang, S.~Zhou, S.~Yu, S.~Zhou, S.~Zheng, T.~Wang, T.~Pei, T.~Yuan, T.~Sun, W.~L. Xiao, W.~Zeng, W.~An, W.~Liu, W.~Liang, W.~Gao, W.~Zhang, X.~Q. Li, X.~Jin, X.~Wang, X.~Bi, X.~Liu, X.~Wang, X.~Shen, X.~Chen, X.~Chen, X.~Nie, X.~Sun, X.~Wang, X.~Liu, X.~Xie, X.~Yu, X.~Song, X.~Zhou, X.~Yang, X.~Lu, X.~Su, Y.~Wu, Y.~K. Li, Y.~X. Wei, Y.~X. Zhu, Y.~Xu, Y.~Huang, Y.~Li, Y.~Zhao, Y.~Sun, Y.~Li,
  Y.~Wang, Y.~Zheng, Y.~Zhang, Y.~Xiong, Y.~Zhao, Y.~He, Y.~Tang, Y.~Piao, Y.~Dong, Y.~Tan, Y.~Liu, Y.~Wang, Y.~Guo, Y.~Zhu, Y.~Wang, Y.~Zou, Y.~Zha, Y.~Ma, Y.~Yan, Y.~You, Y.~Liu, Z.~Z. Ren, Z.~Ren, Z.~Sha, Z.~Fu, Z.~Huang, Z.~Zhang, Z.~Xie, Z.~Hao, Z.~Shao, Z.~Wen, Z.~Xu, Z.~Zhang, Z.~Li, Z.~Wang, Z.~Gu, Z.~Li, and Z.~Xie, ``Deepseek-v2: A strong, economical, and efficient mixture-of-experts language model,'' 2024. [Online]. Available: \url{https://arxiv.org/abs/2405.04434}
\BIBentrySTDinterwordspacing

\bibitem{deepseekv3}
\BIBentryALTinterwordspacing
DeepSeek-AI, A.~Liu, B.~Feng, B.~Xue, B.~Wang, B.~Wu, C.~Lu, C.~Zhao, C.~Deng, C.~Zhang, C.~Ruan, D.~Dai, D.~Guo, D.~Yang, D.~Chen, D.~Ji, E.~Li, F.~Lin, F.~Dai, F.~Luo, G.~Hao, G.~Chen, G.~Li, H.~Zhang, H.~Bao, H.~Xu, H.~Wang, H.~Zhang, H.~Ding, H.~Xin, H.~Gao, H.~Li, H.~Qu, J.~L. Cai, J.~Liang, J.~Guo, J.~Ni, J.~Li, J.~Wang, J.~Chen, J.~Chen, J.~Yuan, J.~Qiu, J.~Li, J.~Song, K.~Dong, K.~Hu, K.~Gao, K.~Guan, K.~Huang, K.~Yu, L.~Wang, L.~Zhang, L.~Xu, L.~Xia, L.~Zhao, L.~Wang, L.~Zhang, M.~Li, M.~Wang, M.~Zhang, M.~Zhang, M.~Tang, M.~Li, N.~Tian, P.~Huang, P.~Wang, P.~Zhang, Q.~Wang, Q.~Zhu, Q.~Chen, Q.~Du, R.~J. Chen, R.~L. Jin, R.~Ge, R.~Zhang, R.~Pan, R.~Wang, R.~Xu, R.~Zhang, R.~Chen, S.~S. Li, S.~Lu, S.~Zhou, S.~Chen, S.~Wu, S.~Ye, S.~Ye, S.~Ma, S.~Wang, S.~Zhou, S.~Yu, S.~Zhou, S.~Pan, T.~Wang, T.~Yun, T.~Pei, T.~Sun, W.~L. Xiao, W.~Zeng, W.~Zhao, W.~An, W.~Liu, W.~Liang, W.~Gao, W.~Yu, W.~Zhang, X.~Q. Li, X.~Jin, X.~Wang, X.~Bi, X.~Liu, X.~Wang, X.~Shen, X.~Chen, X.~Zhang, X.~Chen, X.~Nie, X.~Sun,
  X.~Wang, X.~Cheng, X.~Liu, X.~Xie, X.~Liu, X.~Yu, X.~Song, X.~Shan, X.~Zhou, X.~Yang, X.~Li, X.~Su, X.~Lin, Y.~K. Li, Y.~Q. Wang, Y.~X. Wei, Y.~X. Zhu, Y.~Zhang, Y.~Xu, Y.~Xu, Y.~Huang, Y.~Li, Y.~Zhao, Y.~Sun, Y.~Li, Y.~Wang, Y.~Yu, Y.~Zheng, Y.~Zhang, Y.~Shi, Y.~Xiong, Y.~He, Y.~Tang, Y.~Piao, Y.~Wang, Y.~Tan, Y.~Ma, Y.~Liu, Y.~Guo, Y.~Wu, Y.~Ou, Y.~Zhu, Y.~Wang, Y.~Gong, Y.~Zou, Y.~He, Y.~Zha, Y.~Xiong, Y.~Ma, Y.~Yan, Y.~Luo, Y.~You, Y.~Liu, Y.~Zhou, Z.~F. Wu, Z.~Z. Ren, Z.~Ren, Z.~Sha, Z.~Fu, Z.~Xu, Z.~Huang, Z.~Zhang, Z.~Xie, Z.~Zhang, Z.~Hao, Z.~Gou, Z.~Ma, Z.~Yan, Z.~Shao, Z.~Xu, Z.~Wu, Z.~Zhang, Z.~Li, Z.~Gu, Z.~Zhu, Z.~Liu, Z.~Li, Z.~Xie, Z.~Song, Z.~Gao, and Z.~Pan, ``Deepseek-v3 technical report,'' 2025. [Online]. Available: \url{https://arxiv.org/abs/2412.19437}
\BIBentrySTDinterwordspacing

\bibitem{Azure}
Microsoft, ``Azure openai service models,'' \url{https://learn.microsoft.com/en-us/azure/cognitive-services/openai/concepts/models}, 2023.

\bibitem{CodeT5+}
\BIBentryALTinterwordspacing
Y.~Wang, H.~Le, A.~D. Gotmare, N.~D.~Q. Bui, J.~Li, and S.~C.~H. Hoi, ``Codet5+: Open code large language models for code understanding and generation,'' \emph{CoRR}, vol. abs/2305.07922, 2023. [Online]. Available: \url{https://doi.org/10.48550/arXiv.2305.07922}
\BIBentrySTDinterwordspacing

\bibitem{deepseekcoder}
\BIBentryALTinterwordspacing
D.~Guo, Q.~Zhu, D.~Yang, Z.~Xie, K.~Dong, W.~Zhang, G.~Chen, X.~Bi, Y.~Wu, Y.~K. Li, F.~Luo, Y.~Xiong, and W.~Liang, ``Deepseek-coder: When the large language model meets programming -- the rise of code intelligence,'' 2024. [Online]. Available: \url{https://arxiv.org/abs/2401.14196}
\BIBentrySTDinterwordspacing

\bibitem{qwen25coder}
\BIBentryALTinterwordspacing
B.~Hui, J.~Yang, Z.~Cui, J.~Yang, D.~Liu, L.~Zhang, T.~Liu, J.~Zhang, B.~Yu, K.~Lu, K.~Dang, Y.~Fan, Y.~Zhang, A.~Yang, R.~Men, F.~Huang, B.~Zheng, Y.~Miao, S.~Quan, Y.~Feng, X.~Ren, X.~Ren, J.~Zhou, and J.~Lin, ``Qwen2.5-coder technical report,'' 2024. [Online]. Available: \url{https://arxiv.org/abs/2409.12186}
\BIBentrySTDinterwordspacing

\bibitem{t5}
C.~Raffel, N.~Shazeer, A.~Roberts, K.~Lee, S.~Narang, M.~Matena, Y.~Zhou, W.~Li, and P.~J. Liu, ``Exploring the limits of transfer learning with a unified text-to-text transformer,'' \emph{J. Mach. Learn. Res.}, vol.~21, pp. 140:1--140:67, 2020.

\bibitem{DBLP:conf/iclr/WeiBZGYLDDL22}
\BIBentryALTinterwordspacing
J.~Wei, M.~Bosma, V.~Y. Zhao, K.~Guu, A.~W. Yu, B.~Lester, N.~Du, A.~M. Dai, and Q.~V. Le, ``Finetuned language models are zero-shot learners,'' in \emph{The Tenth International Conference on Learning Representations, {ICLR} 2022, Virtual Event, April 25-29, 2022}.\hskip 1em plus 0.5em minus 0.4em\relax OpenReview.net, 2022. [Online]. Available: \url{https://openreview.net/forum?id=gEZrGCozdqR}
\BIBentrySTDinterwordspacing

\bibitem{ExT5}
\BIBentryALTinterwordspacing
V.~Aribandi, Y.~Tay, T.~Schuster, J.~Rao, H.~S. Zheng, S.~V. Mehta, H.~Zhuang, V.~Q. Tran, D.~Bahri, J.~Ni, J.~P. Gupta, K.~Hui, S.~Ruder, and D.~Metzler, ``Ext5: Towards extreme multi-task scaling for transfer learning,'' in \emph{The Tenth International Conference on Learning Representations, {ICLR} 2022, Virtual Event, April 25-29, 2022}.\hskip 1em plus 0.5em minus 0.4em\relax OpenReview.net, 2022. [Online]. Available: \url{https://openreview.net/forum?id=Vzh1BFUCiIX}
\BIBentrySTDinterwordspacing

\bibitem{T0}
\BIBentryALTinterwordspacing
V.~Sanh, A.~Webson, C.~Raffel, S.~H. Bach, L.~Sutawika, Z.~Alyafeai, A.~Chaffin, A.~Stiegler, A.~Raja, M.~Dey, M.~S. Bari, C.~Xu, U.~Thakker, S.~S. Sharma, E.~Szczechla, T.~Kim, G.~Chhablani, N.~V. Nayak, D.~Datta, J.~Chang, M.~T. Jiang, H.~Wang, M.~Manica, S.~Shen, Z.~X. Yong, H.~Pandey, R.~Bawden, T.~Wang, T.~Neeraj, J.~Rozen, A.~Sharma, A.~Santilli, T.~F{\'{e}}vry, J.~A. Fries, R.~Teehan, T.~L. Scao, S.~Biderman, L.~Gao, T.~Wolf, and A.~M. Rush, ``Multitask prompted training enables zero-shot task generalization,'' in \emph{The Tenth International Conference on Learning Representations, {ICLR} 2022, Virtual Event, April 25-29, 2022}.\hskip 1em plus 0.5em minus 0.4em\relax OpenReview.net, 2022. [Online]. Available: \url{https://openreview.net/forum?id=9Vrb9D0WI4}
\BIBentrySTDinterwordspacing

\bibitem{UnifiedQA}
\BIBentryALTinterwordspacing
D.~Khashabi, S.~Min, T.~Khot, A.~Sabharwal, O.~Tafjord, P.~Clark, and H.~Hajishirzi, ``Unifiedqa: Crossing format boundaries with a single {QA} system,'' in \emph{Findings of the Association for Computational Linguistics: {EMNLP} 2020, Online Event, 16-20 November 2020}, ser. Findings of {ACL}, T.~Cohn, Y.~He, and Y.~Liu, Eds., vol. {EMNLP} 2020.\hskip 1em plus 0.5em minus 0.4em\relax Association for Computational Linguistics, 2020, pp. 1896--1907. [Online]. Available: \url{https://doi.org/10.18653/v1/2020.findings-emnlp.171}
\BIBentrySTDinterwordspacing

\bibitem{ZeroPrompt}
\BIBentryALTinterwordspacing
H.~Xu, Y.~Chen, Y.~Du, N.~Shao, Y.~Wang, H.~Li, and Z.~Yang, ``Zeroprompt: Scaling prompt-based pretraining to 1, 000 tasks improves zero-shot generalization,'' in \emph{Findings of the Association for Computational Linguistics: {EMNLP} 2022, Abu Dhabi, United Arab Emirates, December 7-11, 2022}, Y.~Goldberg, Z.~Kozareva, and Y.~Zhang, Eds.\hskip 1em plus 0.5em minus 0.4em\relax Association for Computational Linguistics, 2022, pp. 4235--4252. [Online]. Available: \url{https://aclanthology.org/2022.findings-emnlp.312}
\BIBentrySTDinterwordspacing

\bibitem{flan-t5}
\BIBentryALTinterwordspacing
H.~W. Chung, L.~Hou, S.~Longpre, B.~Zoph, Y.~Tay, W.~Fedus, E.~Li, X.~Wang, M.~Dehghani, S.~Brahma, A.~Webson, S.~S. Gu, Z.~Dai, M.~Suzgun, X.~Chen, A.~Chowdhery, S.~Narang, G.~Mishra, A.~Yu, V.~Y. Zhao, Y.~Huang, A.~M. Dai, H.~Yu, S.~Petrov, E.~H. Chi, J.~Dean, J.~Devlin, A.~Roberts, D.~Zhou, Q.~V. Le, and J.~Wei, ``Scaling instruction-finetuned language models,'' \emph{CoRR}, vol. abs/2210.11416, 2022. [Online]. Available: \url{https://doi.org/10.48550/arXiv.2210.11416}
\BIBentrySTDinterwordspacing

\bibitem{DBLP:conf/nips/Ouyang0JAWMZASR22}
L.~Ouyang, J.~Wu, X.~Jiang, D.~Almeida, C.~L. Wainwright, P.~Mishkin, C.~Zhang, S.~Agarwal, K.~Slama, A.~Ray, J.~Schulman, J.~Hilton, F.~Kelton, L.~Miller, M.~Simens, A.~Askell, P.~Welinder, P.~F. Christiano, J.~Leike, and R.~Lowe, ``Training language models to follow instructions with human feedback,'' in \emph{NeurIPS}, 2022.

\bibitem{alpaca}
R.~Taori, I.~Gulrajani, T.~Zhang, Y.~Dubois, X.~Li, C.~Guestrin, P.~Liang, and T.~B. Hashimoto, ``Stanford alpaca: An instruction-following llama model,'' 2023.

\bibitem{wang2022self}
Y.~Wang, Y.~Kordi, S.~Mishra, A.~Liu, N.~A. Smith, D.~Khashabi, and H.~Hajishirzi, ``Self-instruct: Aligning language model with self generated instructions,'' \emph{arXiv preprint arXiv:2212.10560}, 2022.

\bibitem{vicuna2023}
\BIBentryALTinterwordspacing
W.-L. Chiang, Z.~Li, Z.~Lin, Y.~Sheng, Z.~Wu, H.~Zhang, L.~Zheng, S.~Zhuang, Y.~Zhuang, J.~E. Gonzalez, I.~Stoica, and E.~P. Xing, ``Vicuna: An open-source chatbot impressing gpt-4 with 90\%* chatgpt quality,'' March 2023. [Online]. Available: \url{https://vicuna.lmsys.org}
\BIBentrySTDinterwordspacing

\bibitem{xu2023wizardlm}
C.~Xu, Q.~Sun, K.~Zheng, X.~Geng, P.~Zhao, J.~Feng, C.~Tao, and D.~Jiang, ``Wizardlm: Empowering large language models to follow complex instructions,'' \emph{arXiv preprint arXiv:2304.12244}, 2023.

\bibitem{yue2023mammoth}
X.~Yue, X.~Qu, G.~Zhang, Y.~Fu, W.~Huang, H.~Sun, Y.~Su, and W.~Chen, ``Mammoth: Building math generalist models through hybrid instruction tuning,'' \emph{arXiv preprint arXiv:2309.05653}, 2023.

\bibitem{chen2022program}
W.~Chen, X.~Ma, X.~Wang, and W.~W. Cohen, ``Program of thoughts prompting: Disentangling computation from reasoning for numerical reasoning tasks,'' \emph{arXiv preprint arXiv:2211.12588}, 2022.

\bibitem{gunasekar2023textbooks}
S.~Gunasekar, Y.~Zhang, J.~Aneja, C.~C.~T. Mendes, A.~Del~Giorno, S.~Gopi, M.~Javaheripi, P.~Kauffmann, G.~de~Rosa, O.~Saarikivi \emph{et~al.}, ``Textbooks are all you need,'' \emph{arXiv preprint arXiv:2306.11644}, 2023.

\bibitem{haluptzok2023language}
P.~Haluptzok, M.~Bowers, and A.~T. Kalai, ``Language models can teach themselves to program better,'' in \emph{The Eleventh International Conference on Learning Representations}, 2023.

\bibitem{west-etal-2022-symbolic}
P.~West, C.~Bhagavatula, J.~Hessel, J.~Hwang, L.~Jiang, R.~Le~Bras, X.~Lu, S.~Welleck, and Y.~Choi, ``Symbolic knowledge distillation: from general language models to commonsense models,'' in \emph{Proceedings of the 2022 Conference of the North American Chapter of the Association for Computational Linguistics: Human Language Technologies}.\hskip 1em plus 0.5em minus 0.4em\relax Seattle, United States: Association for Computational Linguistics, Jul. 2022, pp. 4602--4625.

\bibitem{sclar2022referee}
M.~Sclar, P.~West, S.~Kumar, Y.~Tsvetkov, and Y.~Choi, ``Referee: Reference-free sentence summarization with sharper controllability through symbolic knowledge distillation,'' \emph{arXiv preprint arXiv:2210.13800}, 2022.

\bibitem{luo2023wizardmath}
H.~Luo, Q.~Sun, C.~Xu, P.~Zhao, J.~Lou, C.~Tao, X.~Geng, Q.~Lin, S.~Chen, and D.~Zhang, ``Wizardmath: Empowering mathematical reasoning for large language models via reinforced evol-instruct,'' \emph{arXiv preprint arXiv:2308.09583}, 2023.

\bibitem{patil2023gorilla}
S.~G. Patil, T.~Zhang, X.~Wang, and J.~E. Gonzalez, ``Gorilla: Large language model connected with massive apis,'' \emph{arXiv preprint arXiv:2305.15334}, 2023.

\bibitem{li-etal-2023-symbolic}
L.~H. Li, J.~Hessel, Y.~Yu, X.~Ren, K.-W. Chang, and Y.~Choi, ``Symbolic chain-of-thought distillation: Small models can also {``}think{''} step-by-step,'' in \emph{Proceedings of the 61st Annual Meeting of the Association for Computational Linguistics (Volume 1: Long Papers)}.\hskip 1em plus 0.5em minus 0.4em\relax Toronto, Canada: Association for Computational Linguistics, Jul. 2023.

\bibitem{zelikman2022star}
E.~Zelikman, Y.~Wu, J.~Mu, and N.~Goodman, ``{ST}ar: Bootstrapping reasoning with reasoning,'' in \emph{Advances in Neural Information Processing Systems}, A.~H. Oh, A.~Agarwal, D.~Belgrave, and K.~Cho, Eds., 2022.

\bibitem{wei2022chain}
J.~Wei, X.~Wang, D.~Schuurmans, M.~Bosma, F.~Xia, E.~Chi, Q.~V. Le, D.~Zhou \emph{et~al.}, ``Chain-of-thought prompting elicits reasoning in large language models,'' \emph{Advances in Neural Information Processing Systems}, vol.~35, pp. 24\,824--24\,837, 2022.

\bibitem{leastmost}
D.~Zhou, N.~Sch{\"{a}}rli, L.~Hou, J.~Wei, N.~Scales, X.~Wang, D.~Schuurmans, O.~Bousquet, Q.~Le, and E.~H. Chi, ``Least-to-most prompting enables complex reasoning in large language models,'' \emph{CoRR}, vol. abs/2205.10625, 2022.

\bibitem{pal}
L.~Gao, A.~Madaan, S.~Zhou, U.~Alon, P.~Liu, Y.~Yang, J.~Callan, and G.~Neubig, ``{PAL:} program-aided language models,'' \emph{CoRR}, vol. abs/2211.10435, 2022.

\bibitem{POT}
W.~Chen, X.~Ma, X.~Wang, and W.~W. Cohen, ``Program of thoughts prompting: Disentangling computation from reasoning for numerical reasoning tasks,'' \emph{CoRR}, vol. abs/2211.12588, 2022.

\bibitem{LewkowyczADDMRS22}
A.~Lewkowycz, A.~Andreassen, D.~Dohan, E.~Dyer, H.~Michalewski, V.~V. Ramasesh, A.~Slone, C.~Anil, I.~Schlag, T.~Gutman{-}Solo, Y.~Wu, B.~Neyshabur, G.~Gur{-}Ari, and V.~Misra, ``Solving quantitative reasoning problems with language models,'' in \emph{NeurIPS}, 2022.

\bibitem{WuJLRSJS22}
Y.~Wu, A.~Q. Jiang, W.~Li, M.~N. Rabe, C.~Staats, M.~Jamnik, and C.~Szegedy, ``Autoformalization with large language models,'' in \emph{NeurIPS}, 2022.

\bibitem{MadaanZ0YN22}
A.~Madaan, S.~Zhou, U.~Alon, Y.~Yang, and G.~Neubig, ``Language models of code are few-shot commonsense learners,'' in \emph{{EMNLP}}.\hskip 1em plus 0.5em minus 0.4em\relax Association for Computational Linguistics, 2022, pp. 1384--1403.

\bibitem{YaoZYDSN023}
S.~Yao, J.~Zhao, D.~Yu, N.~Du, I.~Shafran, K.~R. Narasimhan, and Y.~Cao, ``React: Synergizing reasoning and acting in language models,'' in \emph{ICLR}.\hskip 1em plus 0.5em minus 0.4em\relax OpenReview.net, 2023.

\bibitem{self-planning}
X.~Jiang, Y.~Dong, L.~Wang, F.~Zheng, Q.~Shang, G.~Li, Z.~Jin, and W.~Jiao, ``Self-planning code generation with large language models,'' \emph{ACM Transactions on Software Engineering and Methodology}, 2023.

\bibitem{jain2023llmassisted}
N.~Jain, T.~Zhang, W.-L. Chiang, J.~E. Gonzalez, K.~Sen, and I.~Stoica, ``Llm-assisted code cleaning for training accurate code generators,'' 2023.

\bibitem{hendrycks2021measuring}
D.~Hendrycks, S.~Basart, S.~Kadavath, M.~Mazeika, A.~Arora, E.~Guo, C.~Burns, S.~Puranik, H.~He, D.~Song, and J.~Steinhardt, ``Measuring coding challenge competence with {{APPS}},'' in \emph{{{NeurIPS}} Datasets and Benchmarks}, 2021.

\bibitem{gpt2}
A.~Radford, J.~Wu, R.~Child, D.~Luan, D.~Amodei, and I.~Sutskever, ``Language models are unsupervised multitask learners,'' 2019.

\bibitem{openai2023gpt4}
OpenAI, ``Gpt-4 technical report,'' 2023.

\bibitem{zhang2023self}
K.~Zhang, Z.~Li, J.~Li, G.~Li, and Z.~Jin, ``Self-edit: Fault-aware code editor for code generation,'' \emph{arXiv preprint arXiv:2305.04087}, 2023.

\bibitem{le2022coderl}
H.~Le, Y.~Wang, A.~D. Gotmare, S.~Savarese, and S.~C.~H. Hoi, ``Coderl: Mastering code generation through pretrained models and deep reinforcement learning,'' \emph{Advances in Neural Information Processing Systems}, vol.~35, pp. 21\,314--21\,328, 2022.

\bibitem{olausson2023demystifying}
T.~X. Olausson, J.~P. Inala, C.~Wang, J.~Gao, and A.~Solar-Lezama, ``Demystifying gpt self-repair for code generation,'' \emph{arXiv preprint arXiv:2306.09896}, 2023.

\bibitem{radon}
\BIBentryALTinterwordspacing
M.~Lacchia, ``Radon: A python tool that computes various metrics for python code,'' 2014, version 5.1.0, accessed on [Access Date]. [Online]. Available: \url{https://radon.readthedocs.io/}
\BIBentrySTDinterwordspacing

\end{thebibliography}

\appendices
\begin{IEEEbiography}
[{\includegraphics[height=1.25in,clip,keepaspectratio]{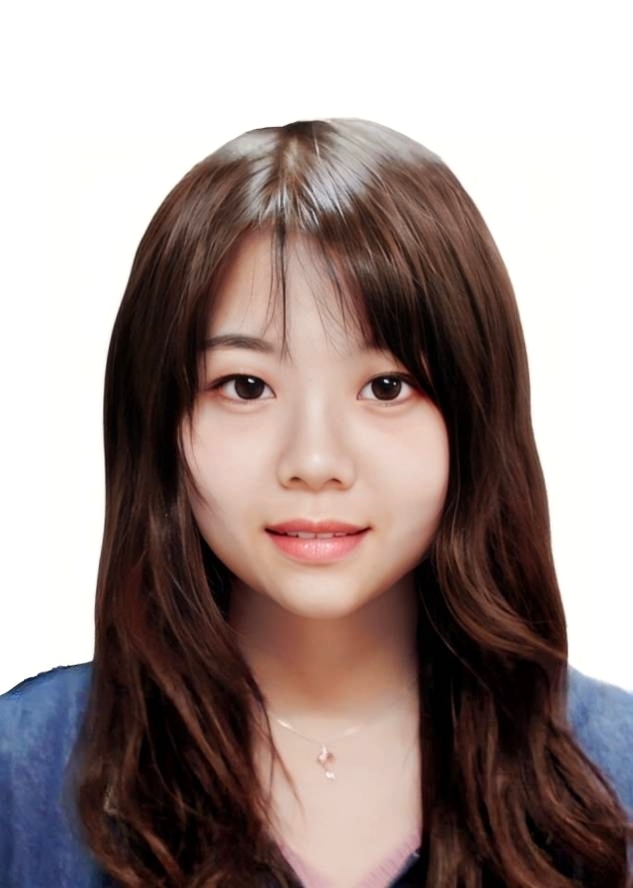}}]
 {Jingyao Li} received the B.Eng. degree from Xi'an Jiaotong University. She is currently a Ph.D. student at Department of Computer Science and Engineering of the Chinese University of Hong Kong (CUHK), under the supervision of Prof. Jiaya Jia. She serves as a reviewer for CVPR, ECCV, ICCV and etc. Her research interests foncus on large language models.
\end{IEEEbiography}
\begin{IEEEbiography}[{\includegraphics[height=1.25in,clip,keepaspectratio]{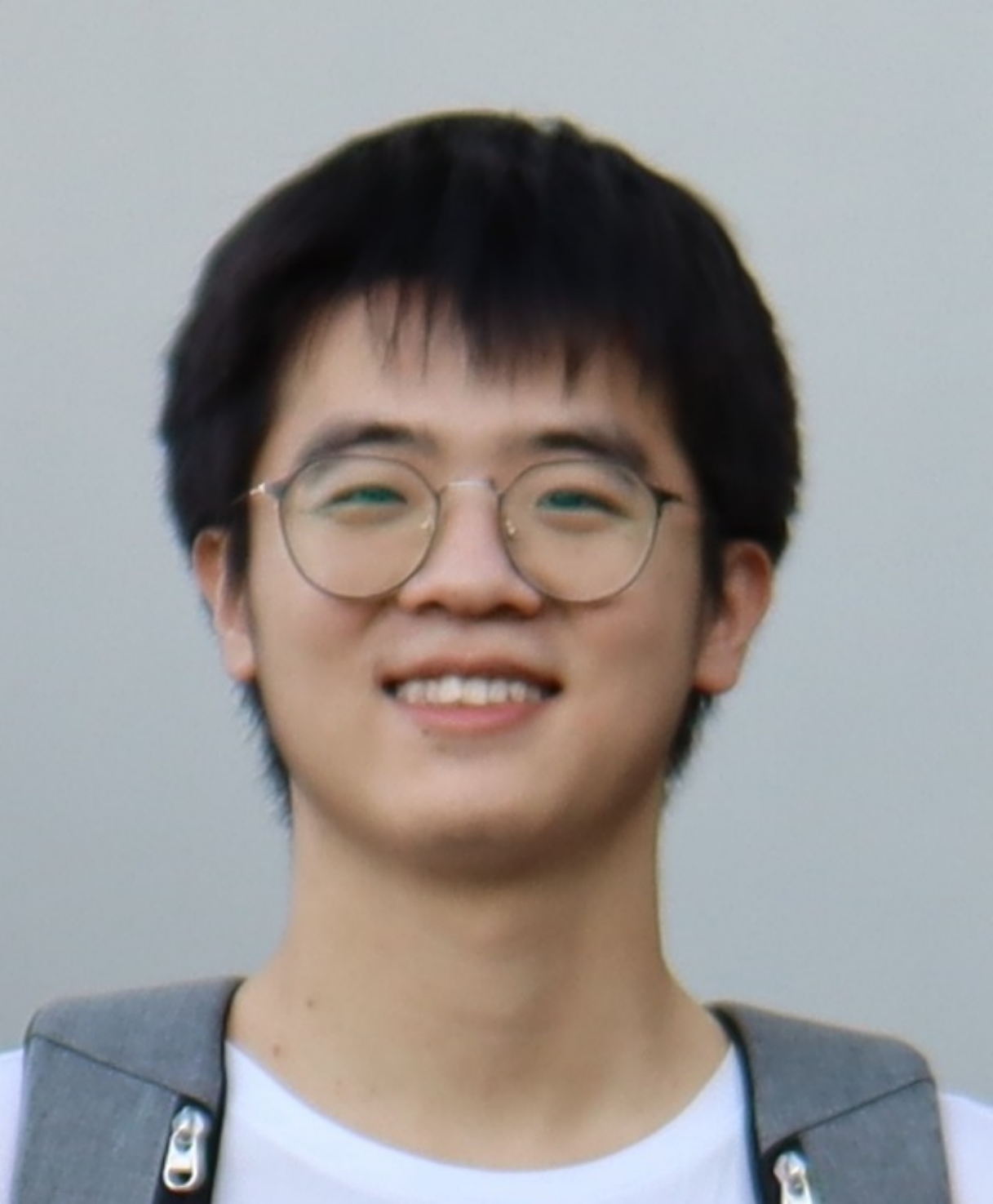}}] {Pengguang Chen} received the B.Eng. degree in Computer Science from Nanjing University and the Ph.D. degree from the Chinese University of Hong Kong (CUHK), under the supervision of Prof. Jiaya Jia. He is currently a researcher in SmartMore. He serves as a reviewer for CVPR, ICCV, ECCV, TPAMI. His research interests include neural architecture search, self-supervised learning, knowledge distillation and semantic segmentation.
\end{IEEEbiography}
\begin{IEEEbiography}[{\includegraphics[width=1in,height=1.25in,clip,keepaspectratio]{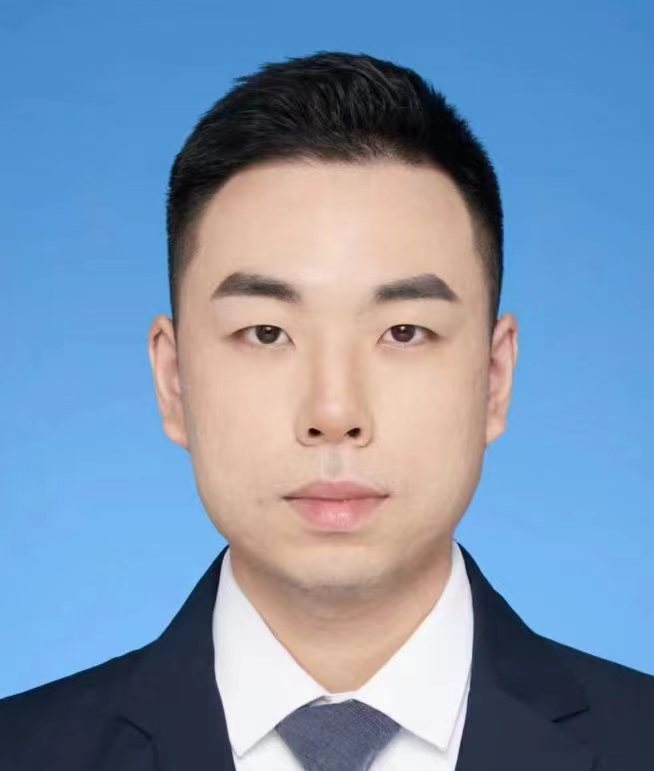}}] {Bin Xia} received the B.E degree in space communication science and technology from Xidian University, Xian, China, in 2020, the M.E degree in communication engineering from Tsinghua University, Beijing, China, in 2020. He is currently pursuing the Ph.D. degree in computer science at Chinese University of Hong Kong, Hong Kong, China. His research interests include image processing, computer vision, large language model, and model compression.
\end{IEEEbiography}
\begin{IEEEbiography}
[{\includegraphics[height=1.25in,clip,keepaspectratio]{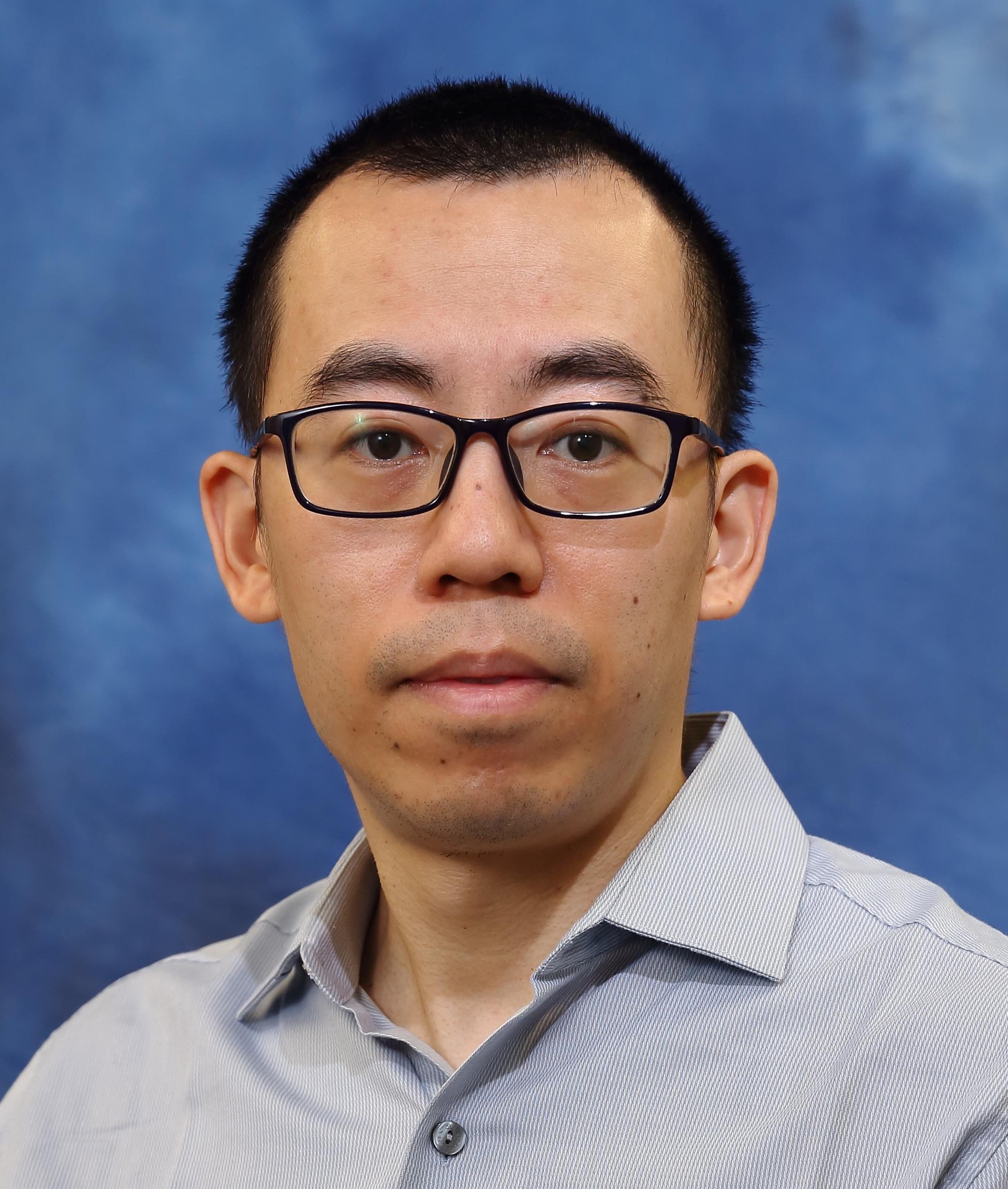}}] 
{Hong Xu} is an Associate Professor in Department of Computer Science and Engineering, The Chinese University of Hong Kong. His research area is computer networking and systems, particularly big data systems and data center networks. From 2013 to 2020 he was with City University of Hong Kong. He received his B.Eng. from The Chinese University of Hong Kong in 2007, and his M.A.Sc. and Ph.D. from University of Toronto in 2009 and 2013, respectively. His work has received best paper awards from ACM SIGCOMM 2022, IEEE ICNP 2023 and 2015, among others. He was the recipient of an Early Career Scheme Grant from the Hong Kong Research Grants Council in 2014. He is a senior member of IEEE and ACM.
\end{IEEEbiography}
\begin{IEEEbiography}
[{\includegraphics[width=1in,height=1.25in,clip,keepaspectratio]{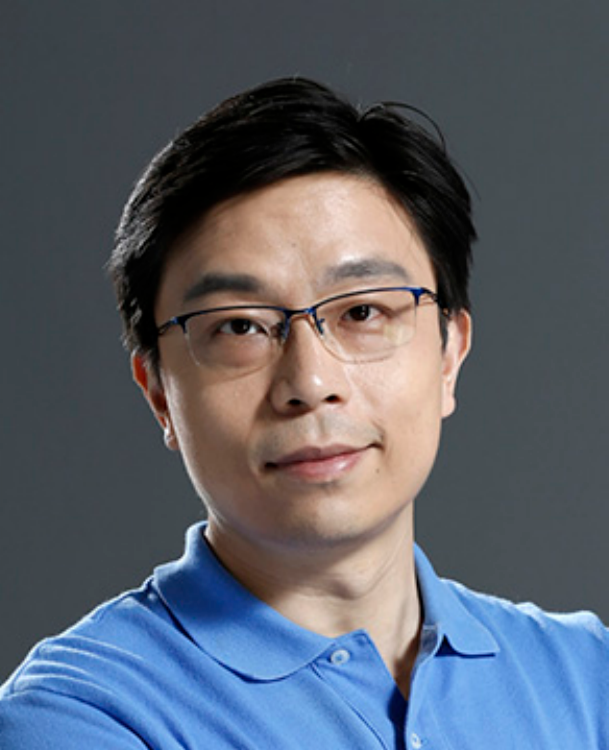}}]{Jiaya Jia} received the Ph.D.~degree in Computer Science from Hong Kong University of Science and Technology in 2004 and is currently a full professor in Department of Computer Science and Engineering at the Chinese University of Hong Kong (CUHK). He assumes the position of Associate Editor-in-Chief of IEEE Transactions on Pattern Analysis and Machine Intelligence (TPAMI) and is in the editorial board of International Journal of Computer Vision (IJCV). He continuously served as area chairs for ICCV, CVPR, AAAI, ECCV, and several other conferences for the organization. He was on program committees of major conferences in graphics and computational imaging, including ICCP, SIGGRAPH, and SIGGRAPH Asia. He is a Fellow of the IEEE. 
\end{IEEEbiography}

\end{document}


\title{\sys: Elevating Large Language Models \\ with \idea}

\author{Jingyao~Li,
        Pengguang~Chen,
        Bin~Xia,
        Hong~Xu,
        and~Jiaya~Jia,~\IEEEmembership{Fellow,~IEEE}
\IEEEcompsocitemizethanks{\IEEEcompsocthanksitem Jingyao Li, Bin Xia and Hong Xu are from the Department of Computer Science and Engineering of the Chinese University of Hong Kong (CUHK) \\
Jiaya Jia's E-mail: leojia9@gmail.com
\IEEEcompsocthanksitem Pengguang Chen and Jiaya Jia are with SmartMore.}
\thanks{Manuscript received Aug. 24th, 2024.}}

\markboth{Journal of \LaTeX\ Class Files,~Vol.~14, No.~8, August~2015}%
{Shell \MakeLowercase{\textit{et al.}}: Bare Demo of IEEEtran.cls for Computer Society Journals}

\maketitle
\IEEEdisplaynontitleabstractindextext
\IEEEpeerreviewmaketitle

\begin{figure*}[h]
    \centering
    \begin{subfigure}[t]{0.49\textwidth}
        \centering
        \includegraphics[trim={0 100 0 0 }, clip, width=\linewidth]{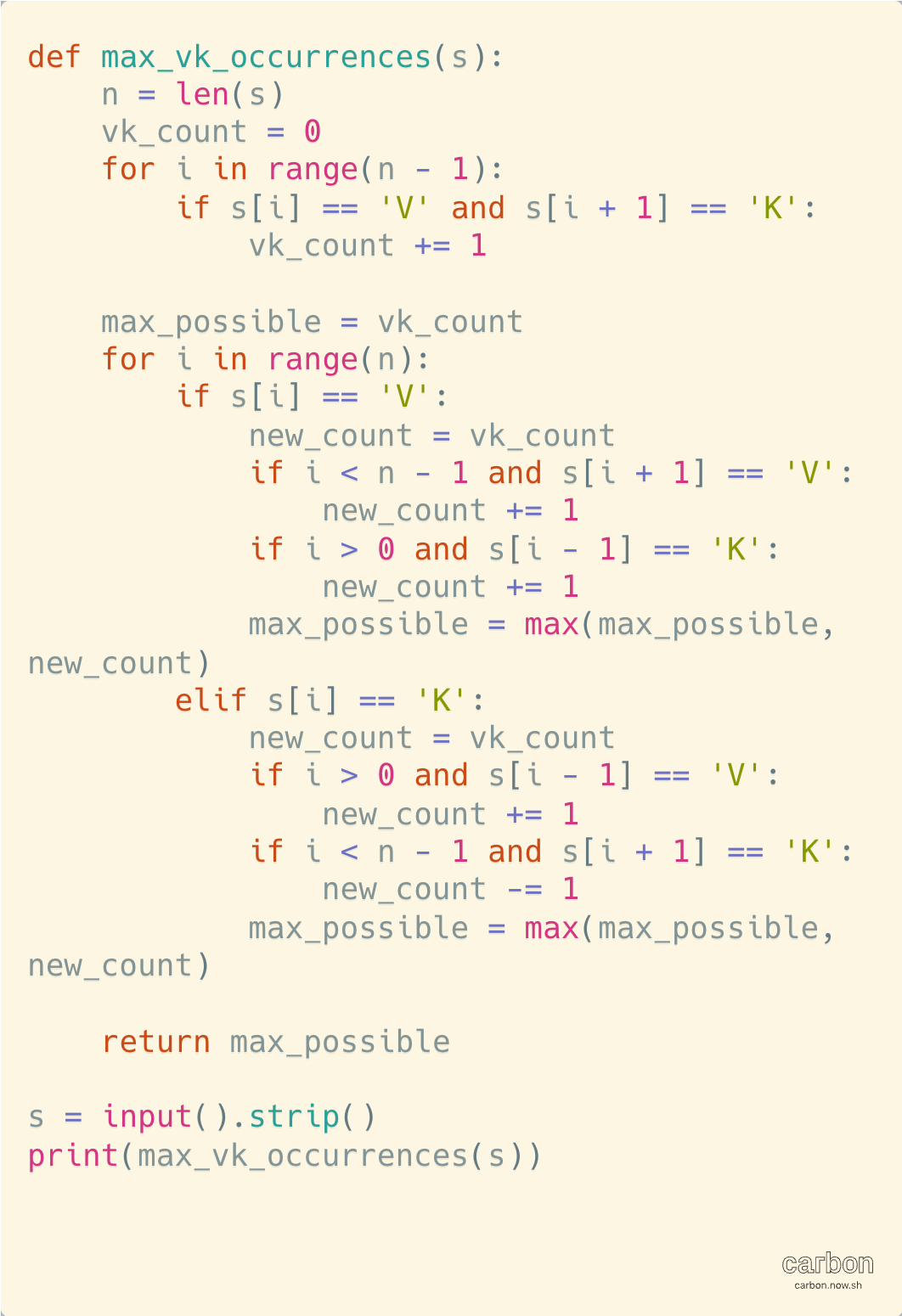} 
        \caption{Qwen2.5-Coder-Instruct-23B: {\color{red}Failed}}
    \end{subfigure}%
    \hfill
    \begin{subfigure}[t]{0.49\textwidth}
        \centering
        \includegraphics[trim={0 100 0 0 }, clip, width=\linewidth]{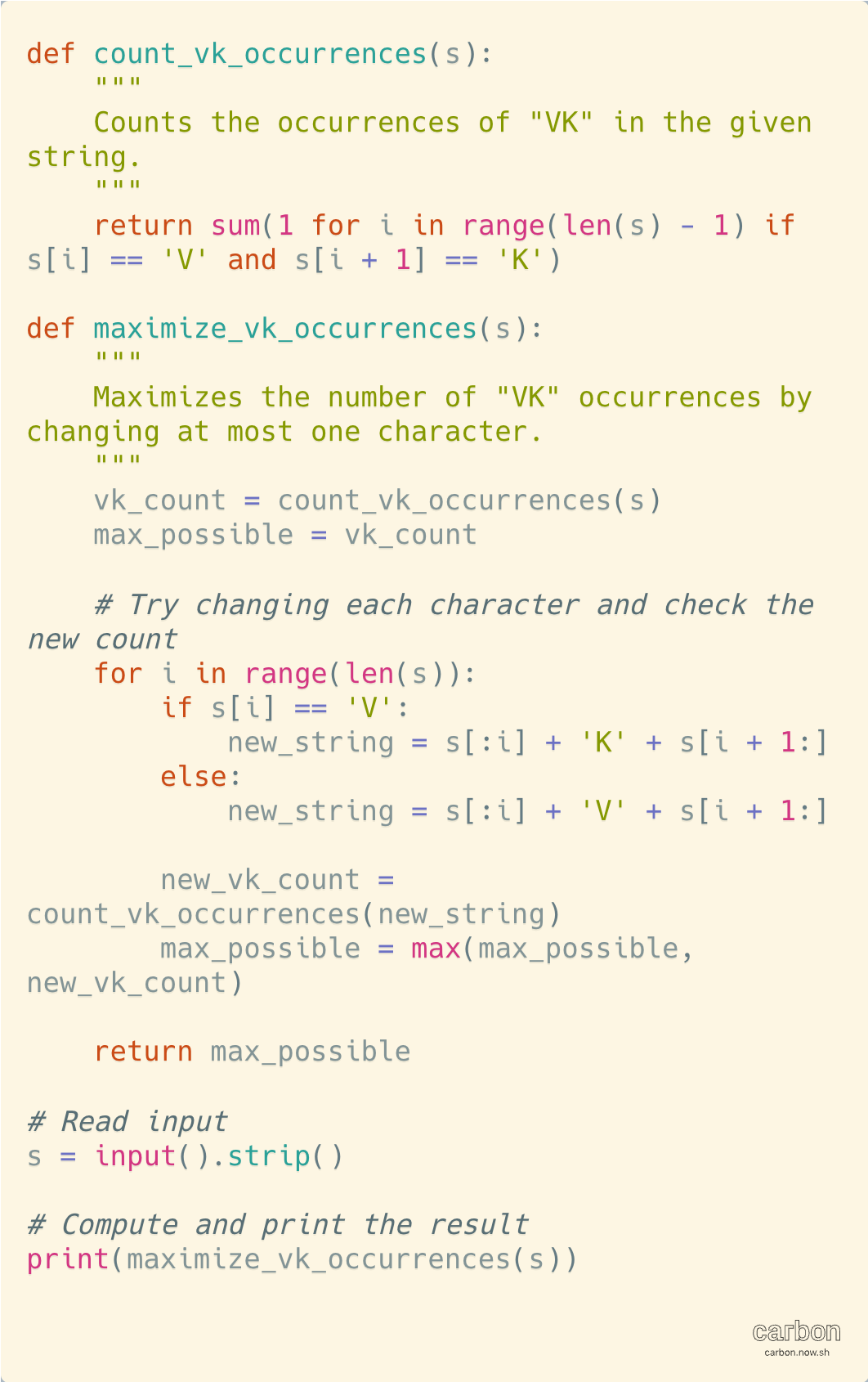}
        \caption{MoTCoder-32B: {\color{teal}Passed}}
    \end{subfigure}
    \caption{Example of codes generated by baseline and our MoTCoder. }
    \label{fig:combined_figures}
\end{figure*}

\begin{figure*}[h]
    \centering
    \begin{subfigure}[t]{\textwidth}
        \centering
        \includegraphics[trim={0 70 0 0 }, clip, width=0.75\linewidth]{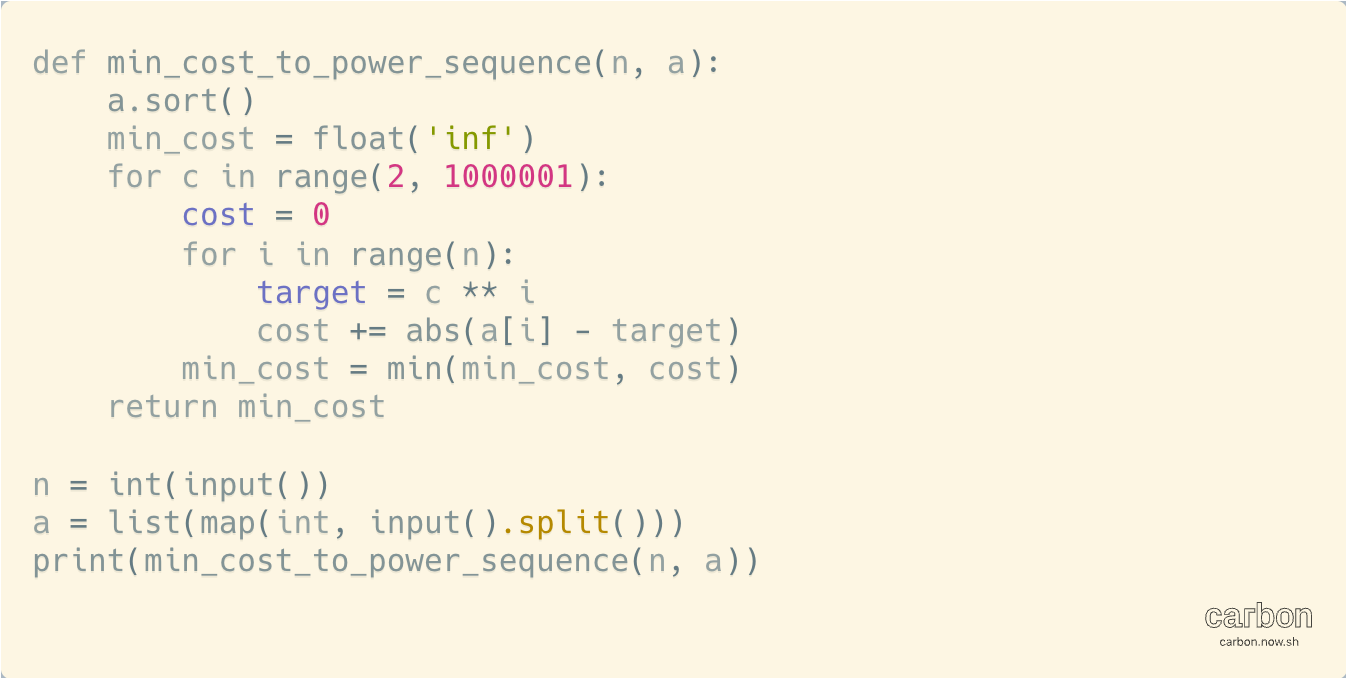} 
        \caption{Qwen2.5-Coder-Instruct-23B: {\color{red}Failed}}
    \end{subfigure}%
    \\
    \begin{subfigure}[t]{\textwidth}
        \centering
        \includegraphics[trim={0 230 350 0 }, clip, width=\linewidth]{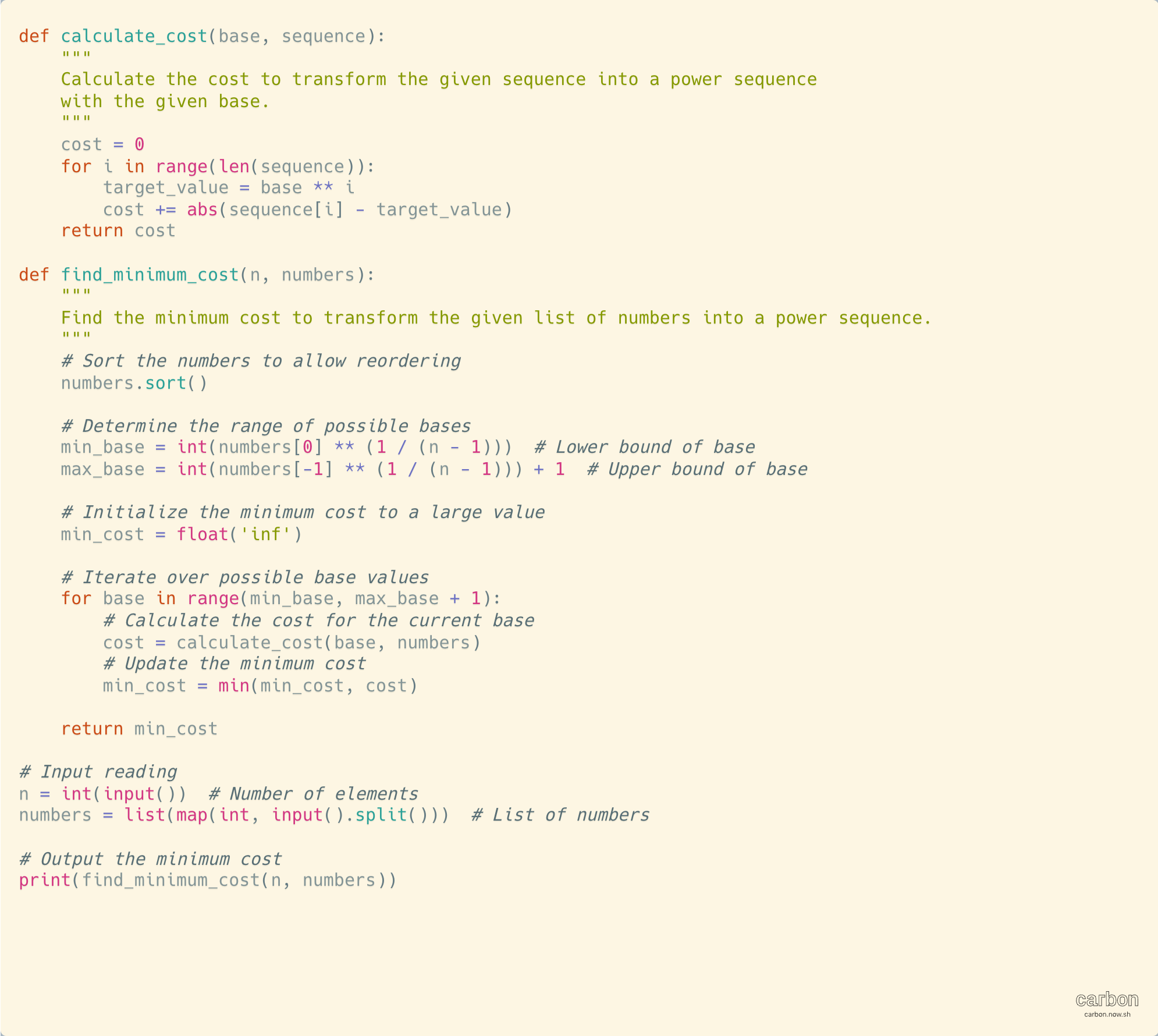}
        \caption{MoTCoder-32B: {\color{teal}Passed}}
    \end{subfigure}
    \caption{Example of codes generated by baseline and our MoTCoder. }
    \label{fig:combined_figures}
\end{figure*}

